\documentclass{article}

\usepackage[final, nonatbib]{neurips_2025}

\usepackage[utf8]{inputenc} %
\usepackage[T1]{fontenc}    %
\usepackage{booktabs}       %
\usepackage{amsfonts}       %
\usepackage{nicefrac}       %
\usepackage{microtype}      %
\usepackage{xcolor}         %
\usepackage{colortbl}
\usepackage{multirow}
\definecolor{fade}{gray}{0.4}
\definecolor{green}{rgb}{0.0, 0.5, 0.0}
\definecolor{grey}{rgb}{0.5, 0.5, 0.5}

\newcommand\fadetext[1]{\textcolor{grey}{#1}}

\usepackage{soul}
\newcommand{\specialcell}[2][c]{%
\begin{tabular}[#1]{@{}c@{}}#2\end{tabular}}
\newcommand{\specialcellleft}[2][l]{%
\begin{tabular}[#1]{@{}l@{}}#2\end{tabular}}

\usepackage{fontawesome}

\usepackage{pifont}%
\usepackage{graphicx}
\usepackage{amsmath}
\usepackage{booktabs}
\usepackage{enumitem}
\usepackage{color, colortbl}
\usepackage{multirow}
\usepackage{diagbox}
\usepackage{algorithm}
\usepackage{listings}
\definecolor{rc1}{RGB}{235,235,235}
\definecolor{rc2}{RGB}{255,255,255}
\definecolor{codeblue}{rgb}{0.25,0.5,0.5}
\definecolor{codekw}{rgb}{0.85, 0.18, 0.50}
\usepackage{booktabs}       %
\usepackage{amsfonts}       %
\usepackage{nicefrac}       %
\usepackage{microtype}      %
\usepackage{graphicx}
\usepackage{tikz}
\usepackage{comment}
\usepackage{amsmath,amssymb} %
\usepackage{color}

\usepackage{stackengine}

\usepackage{algpseudocode}

\definecolor{green}{rgb}{0.0, 0.5, 0.0}
\newcommand\green[1]{{\color{green}{#1}}}

\usepackage{caption}
\usepackage{subcaption}

\usepackage[most]{tcolorbox}
\usepackage{mdframed}
\usepackage{wrapfig}

\usepackage{bbm}
\usepackage{fix-cm} 

\usepackage{xtab}

\usepackage{alltt,xcolor}
\usepackage{relsize}
\usepackage{rotating} 
\usepackage{amsmath,amssymb,amsthm}

\newcounter{qcounter}
\newcommand{\qbold}[1]{\stepcounter{qcounter}\textbf{Q\arabic{qcounter}. #1}}

\usepackage{verbatim}

\newtcblisting[auto counter,number within=chapter]{sourcecode}[2][]{
  sharp corners,
  breakable,
  fonttitle=\bfseries,
  colframe=black!70,
  listing only,
  listing options={
    basicstyle=\ttfamily\scriptsize,
    language=php,
    showstringspaces=false,
    breakatwhitespace=true,
    breaklines=true,
    postbreak=\mbox{},
    tabsize=2,
    linewidth=\linewidth,
    breakindent=0pt
  },
  title=#2,
  #1
}

\newcommand{\greenuline}[1]{%
  \begingroup
  \leavevmode
  \setbox0=\hbox{#1}%
  \begin{tikzpicture}[baseline=(text.base)]
    \node[inner sep=1pt, outer sep=1pt] (text) {\box0};
    \draw[green, thick] (text.south west) -- (text.south east);
  \end{tikzpicture}%
  \endgroup
}

{
    \begin{tcolorbox}[
        width=#1, 
        colframe=black, 
        colback=gray!10, 
        sharp corners, 
        boxrule=0.5mm, 
        left=5pt, right=5pt, top=5pt, bottom=5pt, 
        fontupper=\ttfamily 
    ]
    \small
}
{
    \end{tcolorbox}
}

\usepackage{amsmath,amsfonts,bm}

\def\eqref#1{equation~\ref{#1}}

\def\1{\bm{1}}

\DeclareMathAlphabet{\mathsfit}{\encodingdefault}{\sfdefault}{m}{sl}
\SetMathAlphabet{\mathsfit}{bold}{\encodingdefault}{\sfdefault}{bx}{n}

\usepackage{url}
\usepackage[pagebackref=true,breaklinks=true,colorlinks,bookmarks=false]{hyperref}
\hypersetup{
    colorlinks=true,
    filecolor=magenta,      
    urlcolor=magenta,
}

\usepackage{xspace}
\newcommand{\giticon}[1][1em]{\includegraphics[height=#1]{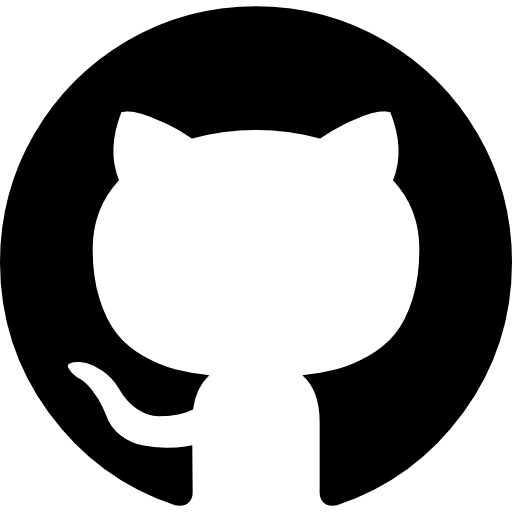}}
\newcommand{\hficon}[1][1em]{\includegraphics[height=#1]{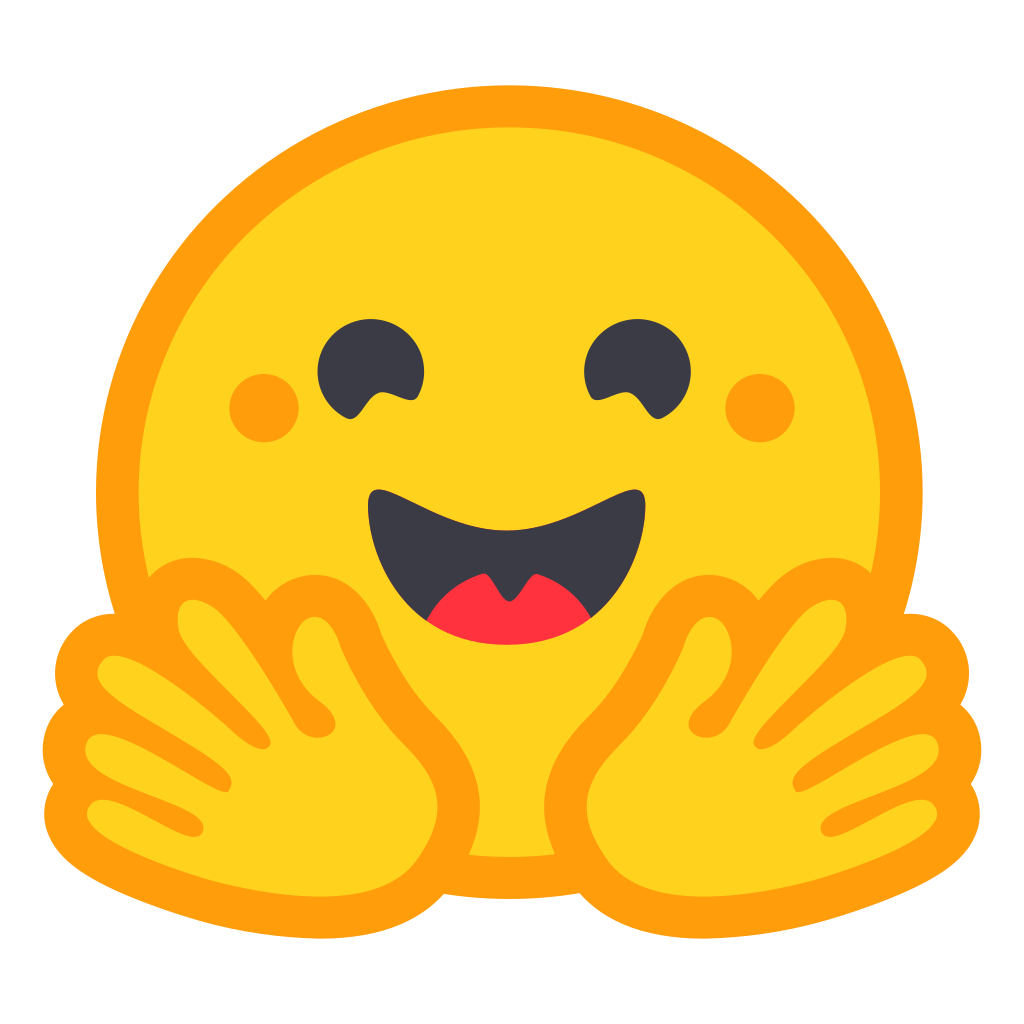}}
\newcommand{\webicon}[1][1em]{\includegraphics[height=#1]{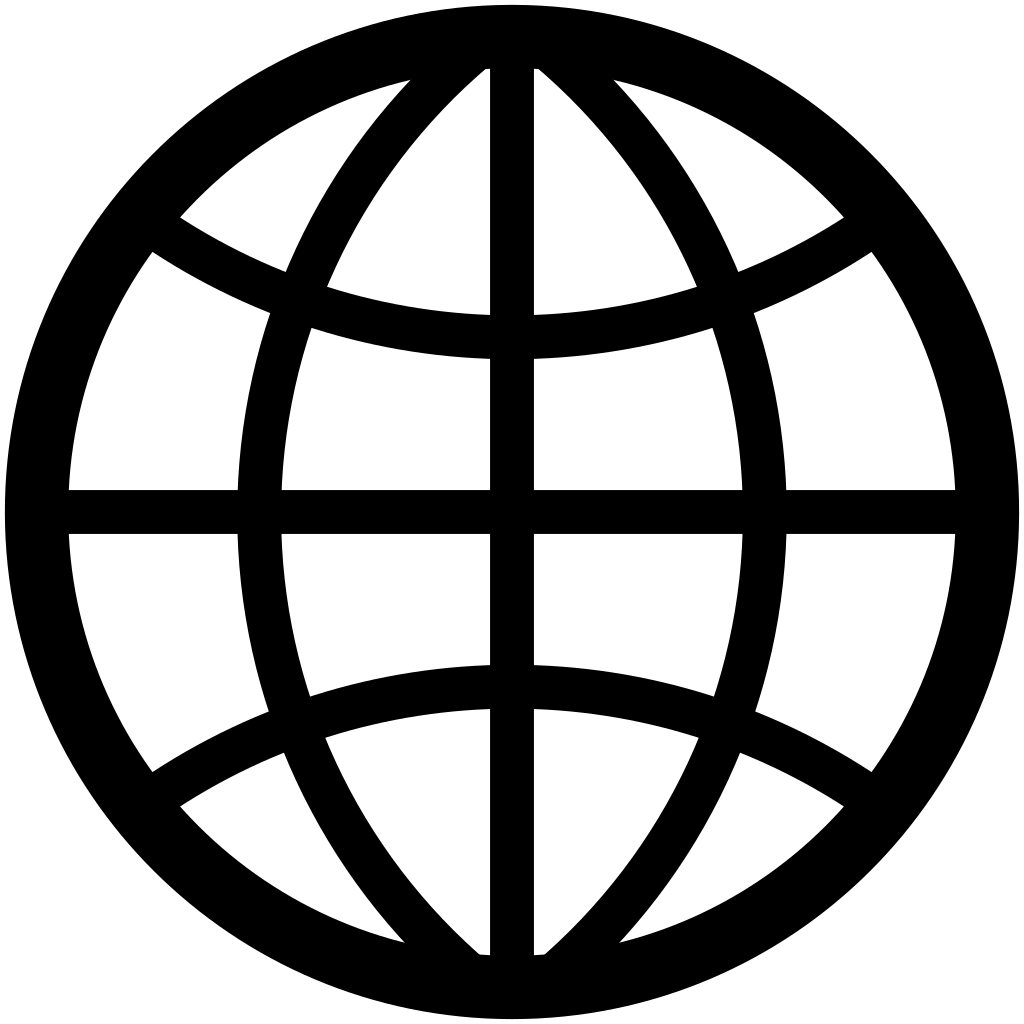}}

\title{
Self-alignment of Large Video Language Models with
\\Refined Regularized Preference Optimization
}

\author{%
Pritam Sarkar \\
Queen's University, Canada and Vector Institute\\
\texttt{pritam.sarkar@queensu.ca} \\
\And
Ali Etemad \\
Queen's University, Canada\\
\texttt{ali.etemad@queensu.ca} \\
\AND
{\begin{tabular}{ccc}
\href{https://pritamqu.github.io/RRPO/}{\webicon[16pt]}
& \href{https://github.com/pritamqu/RRPO}{\giticon[16pt]}
& \href{https://huggingface.co/collections/pritamqu/rrpo-67fbc8c048b298a5fdfb167b}{\hficon[16pt]}
\\
Website & Code & Models \\
\end{tabular}}
}

\begin{document}

\maketitle

\setcounter{footnote}{0}

\begin{abstract}
Despite recent advances in Large Video Language Models (LVLMs), they still struggle with fine-grained temporal understanding, hallucinate, and often make simple mistakes on even simple video question-answering tasks, all of which pose significant challenges to their safe and reliable deployment in real-world applications. To address these limitations, we propose a self-alignment framework that enables LVLMs to learn from their own errors. Our proposed framework first obtains a training set of preferred and non-preferred response pairs, where non-preferred responses are generated by incorporating common error patterns that often occur due to inadequate spatio-temporal understanding, spurious correlations between co-occurring concepts, and over-reliance on linguistic cues while neglecting the vision modality, among others. To facilitate self-alignment of LVLMs with the constructed preferred and non-preferred response pairs, we introduce Refined Regularized Preference Optimization (RRPO), a novel preference optimization method that utilizes sub-sequence-level refined rewards and token-wise KL regularization to address the limitations of Direct Preference Optimization (DPO). We demonstrate that RRPO achieves more precise alignment and more stable training compared to DPO. Our experiments and analysis validate the effectiveness of our approach across diverse video tasks, including video hallucination, short- and long-video understanding, and fine-grained temporal reasoning.

\end{abstract}

\section{Introduction}
\label{sec:intro}

\begin{wraptable}{r}{0.55\textwidth}
\vspace{-0.25in}
\centering
\includegraphics[width=\linewidth]{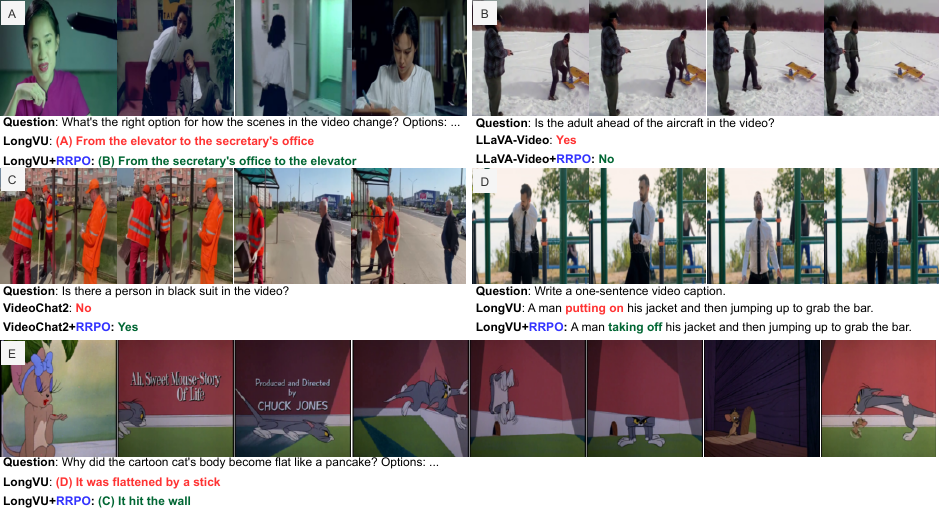}
\captionof{figure}{Examples of failure cases in simple video understanding tasks by state-of-the-art LVLMs, and improvements observed after self-alignment with RRPO.
}
\label{fig:failed_examples}
\vspace{-0.25in}
\end{wraptable}
Despite recent progress in Large Video Language Models (LVLMs) \cite{longvu,slowfast_video,videochat2,llamavid,videollama2,videollama3,timechat,longvila,llavanextvid,videollava,gemini15,internvl,internvl25,qwenvl,qwenvl25},  
these models continue to face limitations in fine-grained temporal understanding \cite{tong2024eyes,tvbench,rahmanzadehgervi2024vision}, demonstrate a propensity for hallucination  \cite{videohallucer,tempcompass}, struggle with long-video understanding \cite{longva,ge2024v2pe,wei2024visual,mlvu,longvideobench}, and frequently make naive mistakes in short-video question-answering tasks \cite{tong2024eyes,tvbench,rahmanzadehgervi2024vision}. 
Please see Figure \ref{fig:failed_examples} for a number of examples. 
These shortcomings severely limit their safe and reliable deployment in real-world applications. The underlying causes of these limitations are multifaceted, encompassing factors such as inadequate spatial and temporal understanding \cite{tong2024eyes,tvbench,rahmanzadehgervi2024vision}, vision-language representational misalignments \cite{shu2025large,ge2024v2pe}, challenges in processing long visual sequences due to context length constraints \cite{longva,ge2024v2pe,wei2024visual}, spurious correlations between co-occurring concepts \cite{dpa,zhou2023analyzing}, and an over-reliance on linguistic cues while neglecting the visual information \cite{tong2024eyes,tvbench,rahmanzadehgervi2024vision,alignment_lvlms}.

\begin{wraptable}{r}{0.55\textwidth}
\vspace{-0.15in}
    \centering
    \includegraphics[width=\linewidth]{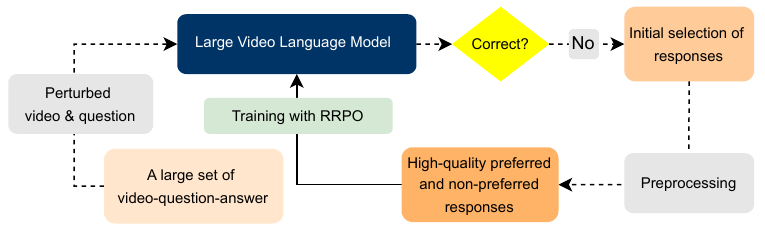}
    \captionof{figure}{An overview of our self-alignment framework.}
    \label{fig:visa_overview}
\vspace{-0.15in}
\end{wraptable}
To address these shortcomings and enhance video understanding in LVLMs, we design a self-alignment \cite{sun2023principle} framework that enables LVLMs to learn from their own errors. Specifically, we begin by sampling video-question pairs from an open-source benchmark. We then apply spatio-temporal perturbations to the video content to mimic common errors that often arise from over-reliance on linguistic cues, spurious correlations between co-occurring concepts, and insufficient spatio-temporal understanding.
Using the perturbed video and the corresponding question, we perform inference with the target LVLM. If the model’s response is incorrect, we construct a self-alignment pair by treating the incorrect response as a non-preferred sample and the correct response as the preferred one, which is kept for self-alignment training. Responses that are already correct are discarded, as they offer limited self-improvement potential. Subsequently, we optimize a loss function to prioritize the preferred response over the non-preferred one. Notably, our data generation pipeline is free from human annotation and can be easily scaled. A high-level overview of our self-alignment framework is depicted in Figure \ref{fig:visa_overview}.

To effectively train LVLMs using the generated preferred and non-preferred response pairs, we introduce \textbf{\textit{Refined Regularized Preference Optimization (RRPO)}}, an approach designed to address limitations of Direct Preference Optimization (DPO) \cite{dpo}. RRPO is designed to address the key drawbacks of DPO. 
RRPO provides a fine-grained sub-sequence-level reward to explicitly penalize the tokens containing the key concepts that are different between the preferred and non-preferred response pairs. This is contrary to DPO's response-level reward which penalizes all tokens within non-preferred responses, thus lacking precision and often proving unsuitable for fine-grained alignment. 
It should be noted that our fine-grained reward further benefits from computing a smaller gradient during optimization and is therefore less prone to diverge away from its initial state unlike DPO's response-level feedback along with weak regularization which can cause significant divergence from the base model, leading to suboptimal performance. However, sub-sequence-level rewards may incentivize the model to exploit shortcuts, such as outputting correct concepts without proper context or complete responses. To mitigate this, we also minimize token-wise KL-divergence \cite{dpa} using a reference model on the preferred responses. This ensures that the LVLM maintains a tight bound on its likelihood across the entirety of the preferred response while reducing the likelihood on the non-preferred concepts.

Through empirical and theoretical analysis, we demonstrate that RRPO exhibits greater stability and smoother convergence during optimization compared to DPO. We validate our method on three popular LVLMs specialized for video understanding, VideoChat2, LLaVA-Video, and longVU, covering a wide range of architectures, LLMs, vision encoders, and training setups. Our in-depth evaluation demonstrates that our proposed self-alignment reduces hallucinations and improves performance in fine-grained temporal understanding, as well as in both short- and long-video understanding tasks.

In summary, our contributions are as follows:
\begin{itemize}
[noitemsep,nolistsep,leftmargin=10pt]
\item 
We design a \textit{self-alignment} framework to facilitate self-improvement of LVLMs based on their own errors.
We introduce \textbf{RRPO}, a preference optimization method that addresses the limitations of DPO by utilizing sub-sequence-level refined rewards and token-wise strong KL regularizer, resulting in more precise alignment and stable training.
\item 
Our rigorous evaluation demonstrates the effectiveness of our proposed method across diverse video tasks, including video hallucination, short and long video understanding, and fine-grained temporal reasoning, among others. Moreover, our experimental and theoretical analyses highlight the superiority of RRPO over DPO in aligning LVLMs. 
We make our code, data, and model weights public to enable fast and accurate reproducibility. 
\end{itemize}

\section{Preference Responses for Self-alignment}

\begin{wraptable}{r}{0.45\linewidth}
\vspace{-0.15in}
    \centering
    \includegraphics[width=\linewidth]{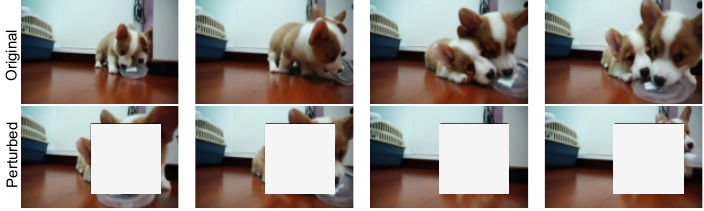}
    \vspace{-0.25in}
    \captionof{figure}{An example of perturbed video.}
    \label{fig:sample_perturb}
\centering
    \vspace{0.1in}
    \includegraphics[width=\linewidth]{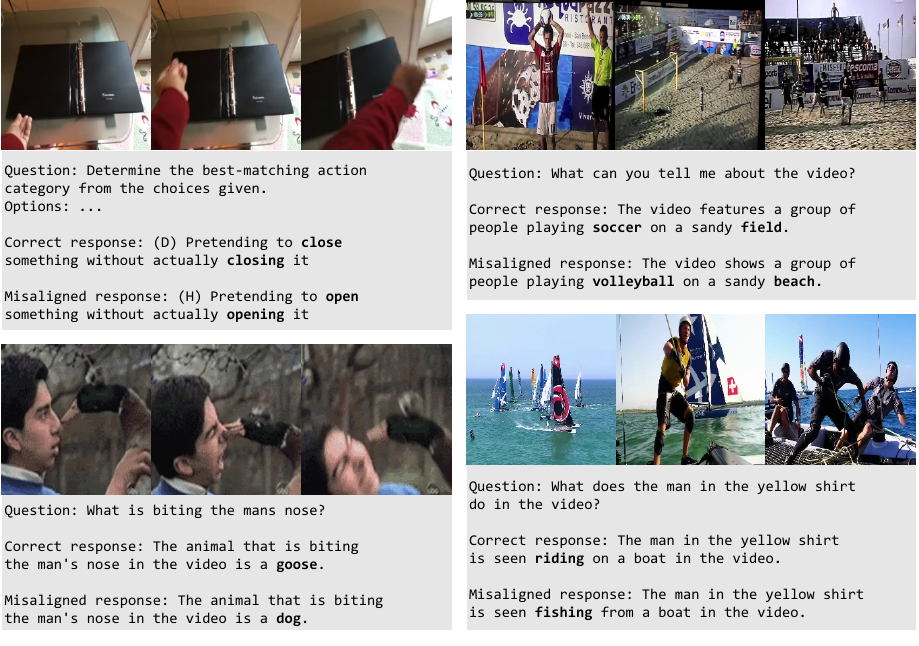}
    \vspace{-0.25in}
    \captionof{figure}{A few training samples. We bold the correct and incorrect concepts.}
    \label{fig:sample_training_data}
\vspace{-0.15in}
\end{wraptable}
As the first step in our framework, we construct a training dataset $\mathcal{D}$ comprising both preferred responses and responses containing incorrect concepts in order to align the LVLM $\pi_\theta$ to favor correct concepts over incorrect ones. 
We begin by utilizing a large and diverse publicly available video instruction tuning dataset, containing triplets of video $x_v$, question $x_q$, and their 
human-preferred answers $y^+$. 
To generate non-preferred responses $y^-$, we obtain perturbed videos $\hat{x}_v\!=\!f_p(x_v)$, where $f_p$ is a perturbation function which masks a large portion of frames and applies temporal shuffling, compromising spatio-temporal consistency. Our intention from this step is to provoke the LVLM to generate responses primarily based on language cues or their parametric knowledge. These perturbed videos $\hat{x}_v$ are fed to $\pi_\theta$ as inputs to generate responses with potential erroneous concepts $y^-\!=\!\pi_\theta(x_q,\hat{x}_v)$.
An example is illustrated in Figure \ref{fig:sample_perturb}. We then verify the correctness of $y^-$, retaining incorrect responses and discarding correct ones. 
Next, We employ an LLM to meticulously compare 
$y^-$ against $y^+$ and identify key incorrect concepts within $y^-$, ensuring that for each incorrect concept $y^-_i\!\in\!y^-$ there exists a corresponding correct concept $y^+_i\!\in\!y^+$. 
In the context of our work, a concept can be actions, objects, attributes, relations, and other key elements in the response that contribute to the semantic understanding of the video.
Furthermore, for lengthy responses, we maintain structural similarity between the incorrect and correct versions by rewriting the correct response while incorporating the incorrect concepts. Finally, we apply deduplication based on these correct-incorrect concept pairs, constructing a high-quality training dataset. Examples are provided in Figure \ref{fig:sample_training_data}.

\section{RRPO}\label{sec:method}

\noindent\textbf{Preliminaries.}
Given an input $x = \{x_q, x_v\}$ with a pair of responses $\{y^+, y^-\}$, where $y^+ \succ y^- | x$, DPO \cite{dpo} can be employed to align $\pi_\theta$ to favor $y^+$ over $y^-$. This is achieved by maximizing the reward margin between $\pi_\theta$ and a reference model $\pi_{\text{ref}}$, using the following training objective:
\begin{equation}\label{eq:dpo}
\mathcal{L}_\text{DPO}(\pi_{\theta};\pi_\text{ref}) = 
-\mathbb{E}_{(x,y^+,y^-) \sim \mathcal{D}} 
\biggl[ \log \sigma \bigl(r_\theta(x,y^+)-r_\theta(x,y^-)\bigr) \biggr],
\end{equation}
where reward $r_\theta(x,y)=\beta\log\frac{\pi_{\theta}(y|x)}{\pi_\text{ref}(y|x)}$ and $\beta$ controls the deviation from $\pi_\text{ref}$. 
Note that $r_\theta(x,y)$ is calculated at a response-level by penalizing all the tokens in $y$, despite the fact that 
the difference between $y^+$ and $y^-$ might only be limited to a \textit{few} key tokens. 
Such response-level reward can be considered coarse-grained reward modeling and is not suitable for fine-grained alignment. Moreover, as the reward is calculated by penalizing all the tokens in the response, the gradient for $\mathcal{L}_\text{DPO}$ tends to be very large for long responses, and accordingly $\pi_\theta$ can diverge to an undesired state thus losing its out-of-the-box capabilities 
\cite{dpa,dpo_issues,wang2024comprehensive}.

\noindent\textbf{Refined Regularized Preference Optimization.}
Our goal is to design a method that can provide a fine-grained reward to penalize individual sub-sequences consisting of the 
tokens belong to the key differing concepts
between $y^+$ and $y^-$. We refer to this as refined reward modeling. 
Let $y$ be expressed as a sequence of tokens $T = \{ t_1, t_2, \ldots, t_{|y|} \}$. Here, the $i$-th sub-sequence $y_{i}$ can be expressed as $T[s_{i} : e_{i}]$, where $s_{i}$ and $ e_{i} $ are the start and end indices of $ y_{i}$ with $ 1 \leq s_{i} \leq e_{i} \leq |y|$. Therefore, the reward for $y_i$ can be computed as:
\begin{equation}\label{eq:subseq_reward}
    r_\theta(x,y_i)=
    \beta\log\frac{
    \prod_{j=s_i}^{e_i} 
    \pi_\theta(t_j|x, t_{<j})
    }{
    \prod_{j=s_i}^{e_i}
    \pi_\text{ref}(t_j|x, t_{<j})
    }.
\end{equation}
Assuming there exists $N$ such sub-sequences, we can train $\pi_\theta$ to maximize the total reward margin 
\begin{equation}\label{eq:total_reward}
u
=\sum\limits_{i=1}^{N}u_i
=\sum\limits_{i=1}^{N}\bigl(r_\theta(x,y^+_i)-r_\theta(x,y^-_i)\bigr).
\end{equation} 
Subsequently, we replace the reward formulation in Equation \ref{eq:dpo} with our refined reward modeling as:
\begin{equation}\label{eq:rank_rrpo}
\mathcal{L}_{\text{RRPO}}^{(\text{rank})}(\pi_{\theta};\pi_\text{ref}) = -\mathbb{E}_{(x,y^+,y^-) \sim \mathcal{D}} \Bigl[ \log \sigma(u) \Bigr].
\end{equation}
However, the sparsity of the rewards used to calculate  $\mathcal{L}_{\text{RRPO}}^{(\text{rank})}$ may allow $\pi_\theta$ to exploit shortcuts and effectively game the reward function by merely outputting key concepts without appropriate context or generating complete responses. 
This \textit{reward hacking} may occur since $\pi_\theta$ is incentivized to maximize reward margin based on the differing sub-sequences; the model can learn to produce short, incomplete responses that contain the correct key concepts, even if those responses lack overall coherence, fluency, or completeness. 
To mitigate this reward hacking, we introduce a regularizer between $\pi_\theta$ and $\pi_{\text{ref}}$, based on token-wise KL (TKL) divergence \cite{dpa,tdpo}, as follows:
\begin{equation}\label{eq:token_kl}
\mathbb{D}_{\text{TKL}} \big(x,y;\pi_\text{ref} \,\|\, \pi_{\theta} \big)=
\sum\limits_{t=1}^{|y|} \mathbb{D}_{\text{KL}} \big( \pi_\text{ref} (\cdot \mid [x, y_{<t}]) \,\|\, \pi_\theta (\cdot \mid [x, y_{<t}]) \big).
\end{equation}
Accordingly, to ensure $\pi_\theta$ retains its original likelihood across the entirety of $y^+$ while reducing the likelihood of non-preferred concepts, we optimize $\mathbb{D}_{\text{TKL}}$ between $\pi_\theta$ and $\pi_\text{ref}$ over $y^+$. We then modify Equation \ref{eq:rank_rrpo} and define the final RRPO loss as:
\begin{equation}\label{eq:rrpo}
\begin{aligned}
\mathcal{L}_\text{RRPO}(\pi_{\theta};\pi_\text{ref})\!=
\!-\mathbb{E}_{(x,y^+,y^-) \sim \mathcal{D}} 
\biggl[ 
\log \sigma (u) 
+ \alpha\!\cdot\!\mathbb{D}_{\text{TKL}} \big(x,y^+\big)
\biggr],
\end{aligned}
\end{equation}
where $\alpha$ controls the divergence between $\pi_\theta$ and $\pi_\text{ref}$, and $\pi_\text{ref}$ is kept fixed. 

\textbf{How is RRPO update different from DPO?}
The gradient for $\mathcal{L}_{\text{RRPO}}$ can be obtained as:
\begin{align}\label{eq:rrpo_grad_final_struct}
\nabla_{\theta} \mathcal{L}_{\text{RRPO}}
\!=\!-\nabla_{\theta}\mathbb{E} 
\biggl[ 
\log \sigma (u) 
+ \alpha\!\cdot\!\mathbb{D}_{\text{TKL}} \big(x,y^+\big)
\biggr]
\!=\!-\mathbb{E}  
\biggl[ 
\nabla_{\theta} \log \sigma (u) 
+ \alpha \cdot \nabla_{\theta} \mathbb{D}_{\text{TKL}} \big(x,y^+\big)
\biggr].
\end{align}
First, let's calculate the gradient for the ranking loss. Using chain rule, we can write 
$\nabla_{\theta} \log \sigma (u)=\Bigl[ \frac{\sigma'(u)}{\sigma(u)} \, \nabla_{\theta} u \Bigr]$.
Using the identity \(\sigma'(u) = \sigma(u)(1-\sigma(u))\), we have 
$\frac{\sigma'(u)}{\sigma(u)} = 1-\sigma(u) = \sigma(-u)$.
Therefore,
$\nabla_{\theta} \log \sigma (u)=\Bigl[ \sigma(-u) \, \nabla_{\theta} u \Bigr]$.
Recalling our sub-sequence-level reward modeling defined in Equation \ref{eq:total_reward},
we can obtain:
\begin{equation}\label{eq:grad_reward}
\nabla_{\theta} u=\sum\limits_{i=1}^{N}\nabla_{\theta} u_i
=\beta \sum\limits_{i=1}^{N}  
\Bigl(
\sum\limits_{j=s_i^+}^{e_i^+} \nabla_\theta \log \pi_\theta(t^+_j|x, t^+_{<j}) 
- 
\sum\limits_{j=s_i^-}^{e_i^-} \nabla_\theta \log \pi_\theta(t^-_j|x, t^-_{<j})
\Bigr),
\end{equation}
as $\pi_\text{ref}$ is fixed. %
Assuming the norm of the gradient of the log-probability for any single token is bounded, $\|\nabla_\theta \log \pi_\theta(t_j|x,t_{<j})\| \leq M$ for some $M>0$, and that each differentiating sub-sequence $y_i^+$ or $y_i^-$ has an average length $L$, we can further bound the gradient norm of the ranking loss:%
\begin{equation}\label{eq:rrpo_grad_norm}
\|\nabla_{\theta} \mathcal{L}_{\text{RRPO}}^{(\text{rank})}\| \le 
\mathbb{E} \Bigl[ \sigma(-u) \|\nabla_{\theta} u\| \Bigr] \le 
\beta \sum\limits_{i=1}^{N} (M L^+_i + M L^-_i) \approx \beta M (2NL),
\end{equation}
where $L$ is the average length of the sub-sequences ($L \approx L_i^+ \approx L_i^-$).
In contrast, the DPO gradient involves sums over the \textit{entire} lengths of $y^+$ and $y^-$. A similar analysis for DPO yields a bound proportional to the total lengths:
\begin{equation}\label{eq:dpo_grad_norm}
\|\nabla_{\theta} \mathcal{L}_{\text{DPO}}\| \le \beta M (|y^+|+|y^-|).
\end{equation}
Crucially, the total length of the differentiating sub-sequences is typically much smaller than the total length of the full responses, i.e., $2NL \ll |y^+| + |y^-|$. Therefore, the upper bound on the gradient magnitude stemming from the ranking objective is smaller for RRPO compared to DPO:
\[
\|\nabla_{\theta} \mathcal{L}_{\text{RRPO}}^{(\text{rank})}\| \ll \|\nabla_{\theta} \mathcal{L}_{\text{DPO}}\|.
\]
Now, let's consider the term $\nabla_{\theta}\mathbb{D}_{\text{TKL}}(\cdot)$ in Equation \ref{eq:rrpo_grad_final_struct}. Based on the formulation of $\mathbb{D}_\text{KL}(\pi_\text{ref}||\pi_\theta)\!=\!\sum_a \pi_\text{ref}(a) \log \frac{\pi_\text{ref}(a)}{\pi_\theta(a)}$ where $a$ represents a token in the vocabulary, and since  $\pi_\text{ref}$ fixed, we derive:
\begin{equation}\label{eq:grad_token_kl}
\nabla_{\theta}\mathbb{D}_{\text{TKL}}(x,y^+)
=\sum\limits_{t=1}^{|y^+|}\nabla_{\theta}\mathbb{D}_{\text{KL}}(\cdot)
=-\sum\limits_{t=1}^{|y^+|}\sum_{a} \pi_{\text{ref}}(a \mid x,y^+_{<t})\, \nabla_\theta \log \pi_\theta(a \mid x,y^+_{<t}).
\end{equation}
Note that $\nabla_{\theta}\mathbb{D}_{\text{TKL}}$ is always negative. Therefore, for $\alpha>0$ in Equation \ref{eq:rrpo_grad_final_struct}, the gradient magnitude of \( \mathcal{L}_{\text{RRPO}} \) is further reduced than $\mathcal{L}_{\text{RRPO}}^{(\text{rank})}$.
\begin{tcolorbox}[colframe=green!60!black,colback=black!5,boxrule=1pt,top=0pt,bottom=0pt,left=0pt,right=0pt]
\[
\|\nabla_{\theta} \mathcal{L}_{\text{RRPO}}\| < \|\nabla_{\theta} \mathcal{L}_{\text{RRPO}}^{(\text{rank})}\| < \|\nabla_{\theta} \mathcal{L}_{\text{DPO}}\|
\]
\end{tcolorbox}
This mathematical derivation confirms that the proposed RRPO loss function effectively reduces the gradient, as initially hypothesized.
The reduced gradient of RRPO ensures more stable updates compared to DPO in gradient-based optimization while simultaneously enabling precise penalties on specific tokens without the risk of significant divergence. 
Furthermore, the $\mathbb{D}_{\text{TKL}}$ term acts as a trust-region constraint \cite{schulman2015trust}, preventing the model from making large, destabilizing updates. As a result, RRPO allows larger learning rates and yielding smoother convergence in practice. 
We present the pseudocode of RRPO in Appendix \ref{supsec:pseudocode}.

\section{Experiment Setup}

\noindent\textbf{Base models.}
We use VideoChat2$_{\text{7B}}$ \cite{videochat2}, LLaVA-Video$_{\text{7B}}$ (also known as LLaVA-Next-Video$_{\text{7B}}$) \cite{llavanextvid}, and LongVU$_{\text{7B}}$ \cite{longvu} as our base models. These models are carefully selected to evaluate our method across diverse LLM architectures, vision encoders, cross-modal adapters, and training setups. 
For instance, among these models, VideoChat2 employs a video encoder, while the others rely on image encoders. Additionally, LongVU incorporates two vision encoders, whereas the rest use a single vision encoder. VideoChat2 utilizes QFormer \cite{blip2} as its cross-modal adapter, whereas LLaVA-Video employs MLP projection layers. These models further differ in their LLM architectures, context lengths, and training setups, among other aspects.

\noindent\textbf{Training data.}
Based on the availability and diversity of video-language instructions, we use VideoChat-IT \cite{videochat2} as our primary source for training samples.
Specifically, we select a subset of VideoChat-IT encompassing eight video datasets: Kinetics700 \cite{uniformerv2}, Something-Something-v2 \cite{ssv2}, VideoChat \cite{videochat}, VideoChatGPT \cite{videochatgpt}, CLEVRER \cite{clevrr}, NEXTQA \cite{nextqa}, EgoQA \cite{ego4d}, and TGIF \cite{tgif}. These datasets span a range of tasks, including video description, question answering, reasoning, and conversation.
For the perturbation step, we mask a significant portion ($25\%$-$50\%$) of each frame and shuffle the temporal order.
We explore three types of temporal perturbations: (\textit{i}) random shuffling, (\textit{ii}) local shuffling, and (\textit{iii}) global shuffling. In random shuffling, frames are shuffled arbitrarily across time. For local shuffling, frames are initially segmented into smaller chunks, and the frames within each chunk are then shuffled. In global shuffling, the order of these chunks is shuffled, rather than individual frames. 
During inference, based on the LVLMs' input capacity, we utilize a maximum of $16$, $64$, and $100$ frames for VideoChat2, LLaVA-Video, and LongVU, respectively. %

\begin{wraptable}{r}{0.3\textwidth}
\vspace{-0.15in}
\centering
    \fontsize{9pt}{10pt}\selectfont
    \setlength\tabcolsep{2pt}
    \caption{Key statistics of the generated training samples. 
    }
    \label{tab:training_data_stat}
    \resizebox{\linewidth}{!}{
    \begin{tabular}{lcc}
    \toprule
    \multirow{1}{*}{\textbf{Model}} & \multirow{1}{*}{\textbf{\# Samples}} & 
    \specialcell{\bf Unique\\\bf pairs}
     \\
    \midrule\midrule
    VideoChat2$_{\text{7B}}$ & $25$K & $18$K  \\
    LongVU$_{\text{7B}}$ & $22$K & $14$K \\
    LLaVA-Video$_{\text{7B}}$ & $21$K & $16$K \\
    \bottomrule
\end{tabular}
}
\vspace{-0.15in}
\end{wraptable}
Following the generation of the responses, we verify their correctness. For Multiple Choice Questions (MCQ) and Binary Question Answering (BinaryQA) tasks, verification is straightforward, using regex-based checks. However, for open-ended questions, this method proves inadequate, as semantically equivalent responses can be expressed in diverse phrasings. Consequently, for open-ended questions, we employ a powerful LLM, GPT-4o-mini \cite{gpt4o}, as a judge \cite{bubeck2023sparks,peng2023instruction,zheng2023judging,chiang2023can}, to ascertain correctness by comparing the generated response with the ground truth from the video instruction tuning dataset. 
Additionally, for long responses, we employ GPT-4o-mini to rewrite the correct response while incorporating the incorrect concepts from the generated response. This ensures that correct and incorrect concepts are aligned across both preferred and non-preferred responses.
The key statistics of our training data are presented in Table \ref{tab:training_data_stat}. The prompts used during pre-processing with GPT-4o-mini are in Appendix \ref{supsec:details_data}.

\noindent\textbf{Implementation details.} 
For all base models, we utilize LoRA \cite{lora} for training, applying it specifically to the LLMs while freezing all other parameters. Unless otherwise specified, we utilize $16$ frames for self-alignment training; the rest follows the default training setup of each respective base model. We use $4\times$ A100 80GB GPUs for training, with the training time varying between $1$ to $10$ hours. Additional implementation details and hyperparameters are provided in Appendix \ref{supsec:implementation_details}.

\noindent\textbf{Evaluation benchmarks.}
To assess the impact of our self-alignment framework, we conduct evaluations across a diverse range of video understanding tasks. Specifically, we choose TVBench \cite{tvbench} and TempCompass \cite{tempcompass} for fine-grained temporal understanding, VideoHallucer \cite{videohallucer} and VidHalluc \cite{vidhalluc} for video hallucination, MVBench \cite{videochat2} and VideoMME \cite{videomme} for short video understanding, and MLVU \cite{mlvu} and LongVideoBench \cite{longvideobench} for long video understanding. 
It should be noted that these benchmarks, while selected for specific tasks, often assess overlapping video understanding capabilities. For instance, while VideoHallucer and VidHalluc are primarily used for hallucination detection, they also evaluate temporal grounding \cite{videohallucer,vidhalluc}. Similarly, while TVBench mainly focuses on temporal understanding, it covers short video understanding as well \cite{tvbench}. Given its inclusion of short, medium, and long videos, VideoMME explores video understanding of varying lengths.

\section{Results and Analysis} \label{sec:results}

This section details our experimental evaluation, which encompasses a comprehensive and comparative analysis against other alignment methods and existing off-the-shelf aligned LVLMs. Furthermore, we present a detailed analysis focusing on the trade-offs between post-alignment divergence and performance. To gain deeper insights into the impact of our proposed method, 
we conduct experiments exploring its effects on fine-grained temporal understanding, hallucination mitigation, and performance on comprehensive video understanding of varying lengths. Finally, our evaluation includes an investigation into the influence of data, the scaling of input frames, and the presence of subtitles on the performance of our method. We conclude with a discussion regarding the generalization capabilities of our approach.

\begin{table*}[!h]
\centering
\begin{minipage}{0.48\textwidth}
    \centering
    \fontsize{9pt}{10pt}\selectfont
    \setlength\tabcolsep{2pt}
    \caption{Comparison with existing preference optimization methods and RRPO ablation variants. 
    $\Delta\!=\!\frac{1}{N}\sum (\text{acc}_\text{aligned}\! - \!\text{acc}_\text{base})$
    and $\%\Delta\!=\!\frac{100}{N}\sum (\text{acc}_\text{aligned}\! - \!\text{acc}_\text{base})/\text{acc}_\text{base}$, where $N$ is the number of evaluation benchmarks used for each ablation.
    RRPO consistently outperforms existing alignment methods.
    }
    \label{tab:ablation_loss}
    \vspace{-0.05in}
    \resizebox{\linewidth}{!}{
    \begin{tabular}{lccccr}
        \toprule
        & \bf TVB & \bf VHall & \bf VMME & \bf MLVU & \bf $\Delta/\%\Delta$
        \\
        \midrule \midrule
        \rowcolor{green!10}
        \bf LongVU$_\text{7B}$ (base) & $53.7$	& $39.2$	& $56.2$	& $63.6$	& $-$ \\ 
        \midrule
        ~~+ DPO \cite{dpo} & $54.3$	& $40.9$	& $56.6$	& $63.6$	& $0.7/1.5$\\
        ~~+ DPA \cite{dpa} & $54.6$	& $40.3$	& $56.9$	& $63.9$	& $0.7/1.5$\\
        ~~+ TDPO \cite{tdpo} & $53.9$	& $41.4$	& $57.0$	& $63.8$	& $0.8/1.9$\\
        ~~+ DDPO \cite{rlhfv} & $54.2$	& $41.7$	& $56.7$	& $63.6$	& $0.9/2.0$\\
        \midrule
        ~~+ RRPO w/o R$^*$ & $54.3$	& $43.0$	& $\mathbf{57.8}$	& $\mathbf{64.5}$	& $1.7/3.8$\\
        ~~+ RRPO w/o $\mathbb{D}_{\text{TKL}}$  & $54.9$	& $39.1$	& $57.4$	& $63.9$	& $0.6/1.1$ \\
        \midrule
        \rowcolor{green!10}
        ~~+ \textbf{RRPO (ours)} & $\mathbf{56.5}$	& $\mathbf{44.0}$	& $57.7$	& $\mathbf{64.5}$	& $\mathbf{2.5/5.4}$ \\
        \bottomrule
    \end{tabular}
    }
    \caption*{\tiny{\textit{(
    Abbreviations, 
    TVB: TVBench; VHall: VideoHallucer; VMME: VideoMME)}}}
\end{minipage}
\hspace{2pt}
\begin{minipage}{0.48\textwidth}
    \centering
    \fontsize{9pt}{10pt}\selectfont
    \begin{tabular}{c}
    \includegraphics[width=.84\linewidth]{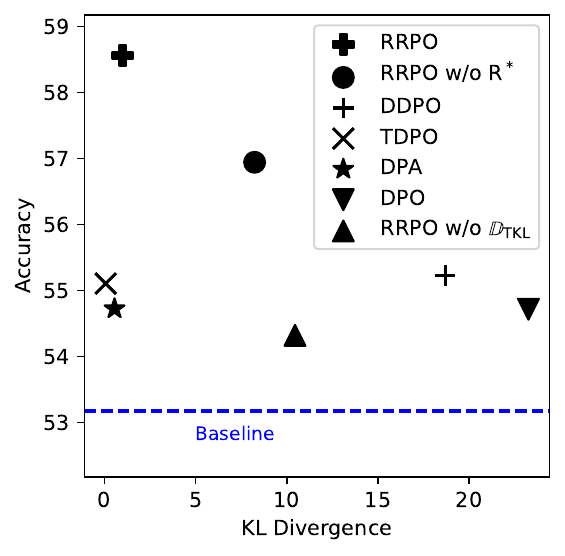}
    \end{tabular}
    \vspace{-0.15in}
    \captionof{figure}{Performance relative to model divergence. RRPO exhibits the best performance-divergence trade-off.}
    \label{fig:score_vs_kld}
\end{minipage}
\vspace{-0.15in}
\end{table*}

\qbold{How well does RRPO perform compared to other alignment methods?}
We conduct an in-depth analysis of RRPO against other recent preference optimization methods designed to address DPO's limitations, namely TDPO \cite{tdpo}, DDPO \cite{rlhfv}, and DPA \cite{dpa} and present the results in Table \ref{tab:ablation_loss}.
More specifically, DDPO was introduced to provide fine-grained rewards, TDPO to enhance DPO's regularization, and DPA to address both of these challenges.
A detailed discussion of their objectives, highlighting similarities and differences, is provided in Appendix \ref{supsec:ablation_variants_details}. 
Furthermore, we include 
two ablation variants of RRPO: one without the refined reward (RRPO w/o R$^*$) and the other without the token-wise KL regularizer (RRPO w/o $\mathbb{D}_{\text{TKL}}$). 
Our study covers various aspects of video understanding, including fine-grained temporal understanding on TVBench, hallucination mitigation on VideoHallucer, comprehensive video understanding on VideoMME, and long video understanding on MLVU. 
The results presented in Table \ref{tab:ablation_loss} demonstrate that RRPO consistently outperforms DPO, DPA, TDPO, and DDPO across all benchmarks. Moreover, among the ablation variants of RRPO, the inclusion of $\mathbb{D}_\text{TKL}$ alone yields significant performance gains, which are further enhanced by incorporating the refined reward. We present qualitative comparisons in Appendix \ref{supsec:rrpo_qualitative}.

\qbold{How much does the model diverge post alignment?}
Understanding the extent of model divergence after alignment is essential to ensure that the process refines behavior without undermining core capabilities as excessive divergence can lead to instability, reduced generalization, and loss of valuable pre-trained knowledge.
Figure \ref{fig:score_vs_kld} presents a comparative analysis of performance improvements against model divergence following preference optimization. 
Despite using a $10\times$ higher learning rate (which is facilitated through the smaller and more stable gradient updates), RRPO exhibits a KL divergence of $1$ compared to DPO's KL divergence of $20$.
While TDPO and DPA exhibit almost the same divergence as RRPO, their performance across all evaluation benchmarks is substantially worse. RRPO, in contrast, demonstrates the optimal performance-divergence trade-off.

\qbold{How does RRPO performance compared to off-the-shelf aligned LVLMs?}
We further evaluate our aligned LVLM against other off-the-shelf aligned LVLMs. Specifically, we compare our RRPO-trained LLaVA-Video with the concurrent work LLaVA-Video-TPO, both of which are based on the LLaVA-Video-Qwen$_\text{7B}$ \cite{llavanextvid}. LLaVA-Video-TPO is trained using a combination of DPO and SFT with a manually curated video-language dataset. The results in Table \ref{tab:compare_concurrent_work} demonstrate that RRPO is significantly more effective than concurrent methods. Our LLaVA-Video-RRPO outperforms LLaVA-Video-TPO across all setups, with performance gains of up to $5.2\%$.

\begin{table}[]%
\centering
    \fontsize{9pt}{10pt}\selectfont
    \setlength\tabcolsep{4pt}
    \caption{Comparison with off-the-shelf aligned LVLMs.
    Ours LLaVA-Video-RRPO outperforms LLaVA-Video-TPO across all setups, both of which are based on LLaVA-Video$_\text{7B}$.
    }
    \label{tab:compare_concurrent_work}
    \resizebox{\linewidth}{!}{
    \begin{tabular}{lcccccccc}
    \toprule
    \bf Model
    & \rotatebox{0}{\specialcellleft{\bf TV\\\bf Bench}} 
    & \rotatebox{0}{\specialcellleft{\bf TempCo-\\\bf mpass$_\text{Avg}$}} 
    & \rotatebox{0}{\specialcellleft{\bf Video\\\bf Hallucer}} 
    & \rotatebox{0}{\specialcellleft{\bf Vid\\\bf Halluc}} 
    & \rotatebox{0}{\specialcellleft{\bf MV\\\bf Bench}} 
    & \rotatebox{0}{\specialcellleft{\bf Video\\\bf MME}} 
    & \rotatebox{0}{\specialcellleft{\bf MLVU$_\text{val}$\\ \bf \small (M-Avg)}} 
    & \rotatebox{0}{\specialcellleft{\bf LongVideo\\\bf Bench$_\text{Val}$}}  \\
    \midrule\midrule
    LLaVA-Video-TPO \cite{llava_tpo} &	$51.1$&	$66.1$&	$50.6$&	$76.3$&	$60.6$&	$\mathbf{65.6}/71.5$&	$68.7$&	$60.1$\\
    LLaVA-Video-RRPO \textbf{(ours)}	& $\mathbf{52.2}$&	$\mathbf{67.4}$&	$\mathbf{55.8}$&	$\mathbf{76.6}$&	$\mathbf{62.0}$&	$65.5/\mathbf{71.8}$&	$\mathbf{69.4}$&	$\mathbf{61.3}$\\
    \bottomrule
    \end{tabular}   
}
\vspace{-0.15in}
\end{table}

\begin{table}[]
\fontsize{9pt}{10pt}\selectfont
\setlength\tabcolsep{4pt}
\centering
\caption{
Evaluating our self-aligned LVLMs on diverse video understanding benchmarks.
We \textbf{bold} the best in each group.
$16$f and $32$f indicate the number of frames utilized during training. 
\#F indicates number of frames used during inference. Here, we presents the overall results averaged across sub-categories where applicable, with detailed results available in Appendix \ref{supsec:addl_results}.
}
\label{tab:main_results}
\resizebox{0.99\linewidth}{!}{
\begin{tabular}{lccccccccc}
\toprule

\bf Models 
& \rotatebox{0}{\specialcellleft{\bf \#F}} 
& \rotatebox{0}{\specialcellleft{\bf TV\\\bf Bench}} 
& \rotatebox{0}{\specialcellleft{\bf TempCo-\\\bf mpass$_\text{Avg}$}} 
& \rotatebox{0}{\specialcellleft{\bf Video\\\bf Hallucer}} 
& \rotatebox{0}{\specialcellleft{\bf Vid\\\bf Halluc}} 
& \rotatebox{0}{\specialcellleft{\bf MV\\\bf Bench}} 
& \rotatebox{0}{\specialcellleft{\bf Video\\\bf MME}} 
& \rotatebox{0}{\specialcellleft{\bf MLVU$_\text{val}$\\ \bf \small (M-Avg)}} 
& \rotatebox{0}{\specialcellleft{\bf LongVideo\\\bf Bench$_\text{Val}$}} 
\\
\midrule \midrule

Video-LLaVA$_\text{7B}$ \cite{videollava} &  & 33.8 & 49.9 & 17.8 & 40.3 & 42.5 & 39.9 & 29.3 & 39.1 \\
VideoLLaMA2$_\text{7B}$ \cite{videollama2} &  & 42.9 & 43.4 & 10.0 & 66.3 & 54.6 & 62.4 & 48.4 & -- \\
LongVA$_\text{7B}$ \cite{longva} &  & -- & 57.0 & -- & -- & -- & 52.6 & 42.1 & -- \\
MiniCPM-V 2.6$_\text{7B}$ \cite{minicpm_v} &  & -- & 66.3 & -- & -- & -- & 60.9 & -- & 54.9 \\
NVILA$_\text{7B}$ \cite{nvila} &  & -- & -- & -- & -- & 68.1 & 64.2 & 70.1 & 57.7 \\
LongVILA$_\text{7B}$ \cite{longvila} &  & -- & -- & -- & -- & 67.1 & 60.1 & -- & 57.1 \\
Qwen2-VL$_\text{7B}$ \cite{qwen2vl} &  & 43.8 & 67.9 & -- & -- & 67.0 & 63.3 & 65.5 & 55.6  \\
LLaVA-NeXT-Video-DPO$_\text{7B}$ \cite{zhang2024llava} &  & -- & 53.8 & 32.0 & -- & -- & -- & -- & 43.5 \\
ShareGPT4Video$_\text{8B}$ \cite{sharegpt4video} &  & -- & 61.5 & 15.8 & 30.9 & -- & 39.9 & 34.2 & 39.7 \\
PLLaVA$_\text{13B}$ \cite{pllava} &  & -- & -- & 41.2 & 48.2 & -- & -- & -- & 45.6 \\
Aria$_\text{8x3.5B}$ \cite{aria} &  & 51.0 & 69.6 & -- & -- & 69.7 & 67.6 & -- & 63.0 \\
LLaVA-NeXT-Video-DPO$_\text{34B}$ \cite{zhang2024llava} &  & -- & -- & 32.3 & 49.5 & -- & -- & -- & 50.5 \\
Qwen2-VL$_\text{72B}$ \cite{qwen2vl} &  & 52.7 & -- & -- & -- & 73.6 & 71.2 & -- & -- \\
GPT-4o \cite{gpt4} &  & 39.9 & 73.8 & 53.3 & 81.2 & 49.1 & 71.9 & 54.5 & 66.7 \\
Gemini 1.5 Pro \cite{gemini15} &  & 47.6 & 67.1 & 37.8 & 72.8 & 60.5 & 75.0 & -- & 64.0 \\

\multicolumn{10}{l}{\textit{\scriptsize (The above models are presented for reference only, and may not be suitable for direct comparisons.)}}
\\
\midrule
\fadetext{VideoChat2$_{\text{7B}}$} \cite{videochat2}
&  & \fadetext{--} & \fadetext{50.8} & \fadetext{7.8} & \fadetext{--} & \fadetext{60.4} & \fadetext{39.5} & \fadetext{--} & \fadetext{39.3} \\
\rowcolor{blue!10}
VideoChat2$_{\text{7B}}$ & $16$
& $44.0$ 	 & $59.3$ 	 & $23.1$ 	 & $73.3$ 	 & $\mathbf{60.2}$ 	 & $41.0$ 	 & $46.4$ 	 & $40.4$ \\
~~+~DPO ($16$f) & $16$
& $45.7$ 	 & $60.0$ 	 & $22.1$ 	 & $72.4$ 	 & $59.6$ 	 & $43.0$ 	 & $47.4$ 	 & $41.0$ \\
\rowcolor{blue!10}
~~+~RRPO ($16$f) (Ours)  & $16$
& $\mathbf{45.8}$ 	 & $\mathbf{60.2}$ 	 & $\mathbf{32.9}$ 	 & $\mathbf{76.4}$ 	 & $59.0$ 	 & $\mathbf{44.3}$ 	 & $\mathbf{47.9}$ 	 & $\mathbf{42.8}$ \\

\midrule
\fadetext{LLaVA-Video$_{\text{7B}}$} \cite{llavanextvid}
&  & \fadetext{45.6} & \fadetext{--} & \fadetext{--} & \fadetext{--} & \fadetext{58.6} & \fadetext{63.3} & \fadetext{70.8} & \fadetext{58.2} \\
\rowcolor{black!10}
LLaVA-Video$_{\text{7B}}$ & $64$
& $51.0$ 	 & $66.0$ 	 & $50.0$ 	 & $\mathbf{76.6}$ 	 & $61.1$ 	 & $64.0$ 	 & $68.6$ 	 & $60.1$ \\
~~+~DPO ($16$f)  & $64$
& $51.9$ 	 & $66.4$ 	 & $53.3$ 	 & $76.5$ 	 & $60.6$ 	 & $63.1$ 	 & $67.4$ 	 & $59.4$ \\
\rowcolor{black!10}
~~+~RRPO ($16$f) (Ours) & $64$
& $51.9$ 	 & $66.8$ 	 & $55.7$ 	 & $76.5$ 	 & $\mathbf{62.2}$ 	 & $\mathbf{64.5}$ 	 & $69.1$ 	 & $\mathbf{60.4}$ \\
\rowcolor{black!10}
~~+~RRPO ($32$f) (Ours) & $64$
& $\mathbf{52.2}$ 	 & $\mathbf{67.4}$ 	 & $\mathbf{55.8}$ 	 & $\mathbf{76.6}$ 	 & ${62.1}$ 	 & $\mathbf{64.5}$ 	 & $\mathbf{69.4}$ 	 & $60.1$ \\

\midrule
\fadetext{LongVU$_{\text{7B}}$} \cite{longvu}
&  & \fadetext{--} & \fadetext{--} & \fadetext{--} & \fadetext{--} & \fadetext{66.9} & \fadetext{--} & \fadetext{65.4} & \fadetext{--} \\
\rowcolor{green!10}
LongVU$_{\text{7B}}$ & $1$fps
& $53.7$ 	 & $63.9$ 	 & $39.2$ 	 & $67.3$ 	 & $65.5$ 	 & $56.2$ 	 & $63.6$ 	 & $48.6$\\
~~+~DPO ($16$f)  & $1$fps
& $54.3$ 	 & $64.3$ 	 & $40.9$ 	 & $68.5$ 	 & $65.9$ 	 & $56.6$ 	 & $63.6$ 	 & $49.4$\\
\rowcolor{green!10}
~~+~RRPO ($16$f) (Ours)  & $1$fps
& $\mathbf{56.5}$ 	 & $\mathbf{64.5}$ 	 & $\mathbf{44.0}$ 	 & $\mathbf{71.7}$ 	 & $\mathbf{66.8}$ 	 & $\mathbf{57.7}$ 	 & $\mathbf{64.5}$ 	 & $\mathbf{49.7}$\\

\bottomrule
\end{tabular}
}
\vspace{-0.1in}
\caption*{\tiny{\textit{
For transparency, we report base LVLMs' results from the literature where available, marked in grey. Differences from our reproduced results, both higher and lower, stem from variations in frame count, input prompt, GPT variant used for evaluation, and occasional implementation issues. For fairness, self-aligned variants are compared against our reproduced results.
}}}
\vspace{-0.15in}
\end{table}

\qbold{Does our method improve fine-grained temporal understanding?}
To evaluate this, we utilize TVBench \cite{tvbench} and TempCompass \cite{tempcompass}, designed to test the 
temporal understanding abilities of LVLMs. TVBench tests capabilities across various temporal tasks, including action localization, directional movement, and scene transitions, among others. Similarly, TempCompass evaluates performance on video captioning, caption matching, MCQ, and BinaryQA, covering video understanding tasks such as event ordering, action identification, and state change. As shown in Table \ref{tab:main_results}, our method, RRPO, consistently improves base model performance by up to $2.8\%$, demonstrating its effectiveness in enhancing fine-grained temporal understanding and outperforming DPO across all setups.

\qbold{Does our method effectively mitigate hallucinations?}
Hallucination occurs when LVLMs produce responses that are ungrounded, referred to as intrinsic hallucination, or unverifiable, referred to as extrinsic hallucination. Hallucination presents a significant obstacle to the reliable use of LVLMs.
To evaluate the impact of our method on video hallucination, we employ VideoHallucer \cite{videohallucer} and VidHalluc \cite{vidhalluc}. 
VideoHallucer tests LVLMs for both extrinsic and intrinsic hallucinations while including both spatial and temporal hallucinations. Additionally, VidHalluc focuses specifically on temporal hallucinations, such as action hallucination. As shown in Table \ref{tab:main_results}, RRPO significantly reduces hallucination across all base models. 
Specifically, RRPO improves performance by $4.8\%$ to $8.8\%$ on VideoHallucer and by up to $4.4\%$ on VidHalluc. In most cases, RRPO demonstrates a substantial performance advantage over DPO, with gains reaching $10.8\%$. 

\qbold{Does our method improve comprehensive video understanding across varying video lengths?}
To evaluate the comprehensive video understanding capabilities of LVLMs across varying video lengths, we leverage four benchmarks: MVBench \cite{videochat2}, VideoMME \cite{videomme}, MLVU \cite{mlvu}, and LongVideoBench \cite{longvideobench}. Together, these benchmarks span a wide variety of perception and reasoning tasks, focusing on objects, actions, their attributes, and holistic video understanding. Among these, MVBench is specifically designed for a comprehensive evaluation of \textit{short videos}, while MLVU and LongVideoBench offer thorough evaluations for \textit{long videos}. VideoMME provides a comprehensive assessment across videos of \textit{varying lengths}.
As shown in Table \ref{tab:main_results}, consistent improvements are observed across all benchmarks for LongVU and LLaVA-Next. For VideoChat2, self-alignment leads to substantial gains in three out of four setups, with only a minor regression in MVBench. Furthermore, RRPO consistently outperforms DPO, further demonstrating the advantages of fine-grained alignment in enhancing comprehensive video understanding.

\begin{table*}[]
\centering
\begin{minipage}{0.48\textwidth} %

    \centering
    \fontsize{9pt}{10pt}\selectfont
    \setlength\tabcolsep{2pt}
    \caption{Impact of different spatio-temporal perturbations in generating non-preferred responses.
    Default perturbations for each model are colored.
    }
    \label{tab:perturbation}
    \vspace{-0.1in}
    \resizebox{0.97\linewidth}{!}{
    \begin{tabular}{lccccr}
        \toprule
         & \bf TVB & \bf VHall & \bf VMME & \bf MLVU & \bf $\Delta/\%\Delta$ \\
        \midrule \midrule

        \rowcolor{blue!10}
        \bf VideoChat2$_\text{7B}$ (base)  & $44.0$ 	 & $23.1$ 	 & $41.0$ 	 & $46.4$ 	 & $-$  \\         \midrule
        ~~+ \texttt{None}  &  $40.7$ 	 & $16.0$ 	 & $39.9$ 	 & $43.8$ 	 & -$3.5/$-$11.6$ \\        \midrule
        ~~+ \texttt{RS}  & $43.0$ 	 & $23.3$ 	 & $41.8$ 	 & $46.3$ 	 & $0.0/0.1$ \\
        ~~+ \texttt{Mask}  & $44.0$ 	 & $26.2$ 	 & $43.4$ 	 & $48.8$ 	 & $2.0/6.1$\\        \midrule
        ~~+ \texttt{LS-Mask}  & $44.6$ 	 & $28.2$ 	 & $44.6$ 	 & $49.4$ 	 & $3.1/9.7$ \\
        ~~+ \texttt{GS-Mask}  & $44.2$ 	 & $28.8$ 	 & $44.1$ 	 & $46.3$ 	 & $2.2/8.1$ \\
        \rowcolor{blue!10}
        ~~+ \texttt{RS-Mask}  & $45.8$ 	 & $32.9$ 	 & $44.3$ 	 & $47.9$ 	 & $\mathbf{4.1/14.4}$ \\
   
        \toprule
        \rowcolor{black!10}
        \bf LLaVA-Video$_\text{7B}$ (base) & $51.0$ 	 & $50.0$ 	 & $64.0$ 	 & $68.6$ 	 & $-$\\ 
        \midrule
        \rowcolor{black!10}
        ~~+ \texttt{LS-Mask}  & $51.9$ 	 & $55.7$ 	 & $64.5$ 	 & $69.1$ 	 & $\mathbf{1.9/3.7}$ \\
        ~~+ \texttt{GS-Mask}  & $51.3$ 	 & $54.8$ 	 & $64.2$ 	 & $68.7$ 	 & $1.4/2.7$ \\
        ~~+ \texttt{RS-Mask}  & $51.5$ 	 & $51.6$ 	 & $64.6$ 	 & $69.6$ 	 & $0.9/1.6$ \\ 
                 
        \toprule
        \rowcolor{green!10}
        \bf LongVU$_\text{7B}$ (base) & $53.7$ 	 & $39.2$ 	 & $56.2$ 	 & $63.6$ 	 & $-$ \\ 
        \midrule
        \rowcolor{green!10}
        ~~+ \texttt{LS-Mask}  & $56.5$ 	 & $44.0$ 	 & $57.7$ 	 & $64.5$ 	 & $\mathbf{2.5/5.4}$ \\
        ~~+ \texttt{GS-Mask}  & $55.0$ 	 & $43.9$ 	 & $56.9$ 	 & $64.5$ 	 & $1.9/4.3$ \\
        ~~+ \texttt{RS-Mask}  & $55.1$ 	 & $42.4$ 	 & $57.0$ 	 & $64.3$ 	 & $1.5/3.3$ \\

        \bottomrule
    \end{tabular}
    }
\end{minipage}%
\hfill
\begin{minipage}{0.48\textwidth} %
    \centering
    \fontsize{9pt}{10pt}\selectfont
    \setlength\tabcolsep{2pt}
    \caption{Impact of data size.}
    \label{tab:data_size}
    \vspace{-0.1in}
    \resizebox{0.8\linewidth}{!}{
    \begin{tabular}{llcccr}
        \toprule
         & \bf TVB & \bf VHall & \bf VMME & \bf MLVU & \bf $\Delta/\%\Delta$ \\
        \midrule \midrule
        Baseline & $51.0$ 	 & $50.0$ 	 & $64.0$ 	 & $68.6$ 	 & $-$ \\ 
        \midrule
        ~~+ $5$K  & $50.9$ 	 & $53.7$ 	 & $64.0$ 	 & $69.0$ 	 & $1.0/1.9$ \\
        ~~+ $10$K  & $51.2$ 	 & $53.8$ 	 & $64.3$ 	 & $69.0$ 	 & $1.2/2.3$ \\
        ~~+ $15$K  & $51.8$ 	 & $54.4$ 	 & $64.2$ 	 & $68.9$ 	 & $1.4/2.8$ \\
        ~~+ $20$K  & $51.9$ 	 & $55.7$ 	 & $64.5$ 	 & $69.1$ 	 & $\mathbf{1.9/3.7}$ \\    
        \bottomrule
    \end{tabular}
    }

    \vspace{5pt} %

    \begin{minipage}{0.48\textwidth} %
        \centering
        \fontsize{9pt}{10pt}\selectfont
        \setlength\tabcolsep{1pt}
        \caption{Impact of varying the number of frames 
        at inference.
        }
        \label{tab:num_frames}
        \vspace{-0.05in}
        \resizebox{\linewidth}{!}{
        \begin{tabular}{llccc}
            \toprule
             & & \bf TVB & \bf MVB & \bf LVB \\
            \midrule \midrule

            \multirow{2}{*}{$32$}  
            & Baseline & $49.6$&    $61.2$& 	$58.4$\\
            & ~~+ RRPO & $51.3$& 	$61.7$& 	$58.9$\\ 
            \midrule
            \multirow{2}{*}{$64$} 
            & Baseline & $51.0$& 	$61.1$& 	$60.1$\\
            & ~~+ RRPO & $\mathbf{52.2}$& 	$\mathbf{62.1}$& 	$60.1$\\         
            \midrule
            \multirow{2}{*}{$128$} 
            & Baseline & $49.4$& 	$60.5$& 	$60.3$\\
            & ~~+ RRPO & $51.3$& 	$61.2$& 	$\mathbf{61.3}$\\ 
            \bottomrule
        \end{tabular}
        }
    \end{minipage}%
    \hfill
    \begin{minipage}{0.48\textwidth} %
        \centering
        \fontsize{9pt}{10pt}\selectfont
        \setlength\tabcolsep{1pt}
        \caption{Impact of using subtitles along with videos (VMME).}
        \label{tab:subtitles}
        \vspace{-0.05in}
        \resizebox{0.93\linewidth}{!}{
        \begin{tabular}{lcc}
            \toprule
             & \bf without & \bf with \\
            \midrule \midrule
            VideoChat2$_\text{7B}$ & $41.0$&$48.0$ \\
            ~~~+ RRPO & $\mathbf{44.3}$ & $\mathbf{49.4}$ \\ 
            \midrule
            LLaVA-Video$_\text{7B}$ & $63.8$ & $67.4$ \\
            ~~~+ RRPO & $\mathbf{64.5}$ & $\mathbf{68.0}$ \\         
            \midrule
            LongVU$_\text{7B}$ & $56.2$ & $62.0$ \\
            ~~~+ RRPO & $\mathbf{57.7}$ & $\mathbf{63.1}$ \\ 
            \bottomrule
        \end{tabular}
        }
    \end{minipage}
\end{minipage}
\vspace{-0.1in}
\caption*{\tiny{\textit{(
Abbreviations, 
\texttt{RS}: Random Shuffle; \texttt{LS}: Local Shuffle; \texttt{GS}: Global Shuffle; 
TVB: TVBench; VHall: VideoHallucer; VMME: VideoMME; LVB: LongVideoBench.)}}}
\vspace{-0.35in}
\end{table*}

\qbold{How do perturbations in our data generation pipeline impact the quality of non-preferred responses?}
To assess the impact of the perturbations 
on the quality of non-preferred responses, we conduct in-depth analyses and present the results in Table \ref{tab:perturbation}. Our key observations are as follows:
First, non-preferred responses generated without video perturbations leads to diminished model performance, likely due to reduced generalizability. Second, temporal perturbations alone are ineffective, although their combination with masking significantly boosts performance. Among the spatio-temporal perturbations, random shuffling with masking (\texttt{RS-Mask}) improves performance for models processing fewer frames (e.g., VideoChat2) whereas local shuffling with masking (\texttt{LS-Mask}) proves superior for models handling longer sequences (e.g., LongVU, LLaVA-Video).

\qbold{How does our method scale with training data?}
To test the impact of scaling the amount of data, we perform an experiment by varying the number of training samples from $5$K to $20$K, incrementing by $5$K. As shown in Table \ref{tab:data_size}, performance improves with data size. This suggests that our data-generation pipeline is effective in producing high-quality training samples for self-alignment.

\qbold{How does performance vary with the number of input frames?}
We investigate the effect of scaling the number of visual input frames during both inference and self-alignment training. 
For inference, we evaluate LLaVA-Video using $32$, $64$, and $128$ frames across TVBench, MVBench, and LongVideoBench. The results are presented in Table \ref{tab:num_frames}. Our key observations are as follows: 
First, RRPO consistently improves performance over the base model across all setups. Second, neither the base model nor RRPO exhibits performance gains beyond $64$ frames on TVBench and MVBench. This is likely due to the short-video nature of these benchmarks, where higher frame counts result in redundant frame resampling. However, for long-video understanding, increasing frame counts yields performance improvements, particularly for RRPO with a $1.2\%$ improvement compared to the base model's $0.2\%$ gain.
Subsequently, we explore the impact of increasing the number of frames during self-alignment training. Specifically, we raise the frame count from our default $16$ to $32$. The results presented in Table \ref{tab:main_results} demonstrate that this increase enhances performance. Notably, we observe a consistent performance improvement on fine-grained temporal understanding tasks, as evidenced by the gains on TVBench and TempCompass.

\qbold{Does our method retain its performance advantage with subtitles?}
Subtitles generally enhance video understanding by providing additional language cues that LVLMs can leverage. Thus, we investigate whether our method maintains its benefits over base models when subtitles are included. As shown in Table \ref{tab:subtitles}, our method demonstrates consistent improvements across all setups.

\qbold{Does our method generalize across diverse LVLM architectures and training setups?} 
Given the rapid evolution of LVLMs, we carefully select VideoChat2, LLaVA-Video, and LongVU as the base models to cover a wide variety of design choices and training methodologies  (e.g., Table \ref{tab:main_results}).
Specifically, VideoChat2 uses the UMT \cite{umt} video encoder, while LLaVA-Video and LongVU use DINOv2 \cite{dinov2} and SigLIP \cite{siglip} as their image encoders. On the other hand, LongVU employs dual vision encoders, unlike the others. 
Moreover, the cross-modal adapters range from query-based for VideoChat2 to MLP projections for LLaVA-Video, and a combinations of both in the case of LongVU. 
Furthermore, the overlap between self-alignment training samples and instruction tuning data differs across models, with VideoChat2 having the highest overlap and LLaVA-Video having the least. Importantly, our self-alignment consistently improves performance across these diverse setups, even when reusing instruction tuning data.%

\section{Related work}
Recent years have seen a surge in the development of LVLMs with improved video understanding capabilities \cite{longvu,slowfast_video,videochat2,llamavid,videollama2,videollama3,timechat,longvila,llavanextvid,videollava,gemini15,internvl,internvl25,qwenvl,qwenvl25}. 
This progress can be attributed to several key factors:
(1) the development of diverse video benchmarks \cite{videochat2,timechat,video_star,ego4d,webvid}, which enable LVLMs to follow human instructions and reason across a variety of video tasks;
(2) architectural innovations that support the use of rich, dense visual features and enable efficient processing of long sequences \cite{longva,longvila,longvu,moviechat,videochat_flash,lita,videollama2,videogptplus}; and
(3) advancements in training algorithms for both the pre-training \cite{internvideo2,umt,siglip,clip} and post-training \cite{rlhf,ppo,dpo} stages.
LVLM training generally involves multi-stage processes \cite{video_llm_survey2,video_llm_survey}, with pre-training typically focusing on representational alignment between video and language \cite{videochat2,llava,blip2}, and post-training refining the model's ability to follow instructions \cite{videochat2,llava,instructblip}, reduce hallucination \cite{rlhfv,dpa}, improve reasoning skills \cite{qwen25vl,zhang2025mm}, and align the model with human preferences \cite{qwen25vl,zhang2025mm}. Our introduced RRPO is a post-training alignment method designed to enhance the overall video understanding capabilities of LVLMs. While concurrent works \cite{videosavi,isrt,llava_tpo} have explored post-training alignments of LVLMs, they directly adopt DPO \cite{dpo}, which proves ineffective in fine-grained alignment, as discussed in Section \ref{sec:results}.

\section{Conclusion}
To improve the video understanding abilities of LVLMs, we design a self-alignment framework that enables LVLMs to learn from their own errors. These errors commonly occur due to their lack of spatio-temporal reasoning, over-reliance on linguistic cues, and spurious correlations between co-occurring concepts, among others. To effectively align LVLMs against such errors, we introduce RRPO, a novel preference optimization method designed for fine-grained alignment through refined reward modeling with strong regularization.
Our in-depth experiments and analyses show that RRPO training is more stable and highly effective compared to prior and concurrent preference optimization methods across diverse tasks. Moreover, the fine-grained reward modeling in RRPO improves capabilities without causing significant divergence from the base models.
Our proposed self-alignment with RRPO exhibits consistent improvements across all setups over the base models, effectively reducing hallucination and improving spatio-temporal reasoning, thus enabling safer and more reliable use of LVLMs. Moreover, we show that the approach scales well with more data and high-resolution temporal inputs, and generalizes well across diverse LVLM architectures and training setups.
Future work can further investigate iterative self-alignment methodologies with RRPO, moving beyond the static dataset used in this work.

\section*{Acknowledgment}
We thank Ahmad Beirami for his valuable feedback and suggestions, which helped improve our paper.
We also thank the Bank of Montreal and Mitacs for funding this research, and the Vector Institute for providing computational resources.

{
\bibliographystyle{unsrt}
\bibliography{egbib}

\begin{thebibliography}{10}

\bibitem{longvu}
Xiaoqian Shen, Yunyang Xiong, Changsheng Zhao, Lemeng Wu, Jun Chen, Chenchen Zhu, Zechun Liu, Fanyi Xiao, Balakrishnan Varadarajan, Florian Bordes, et~al.
\newblock Longvu: Spatiotemporal adaptive compression for long video-language understanding.
\newblock {\em arXiv preprint arXiv:2410.17434}, 2024.

\bibitem{slowfast_video}
Mingze Xu, Mingfei Gao, Zhe Gan, Hong-You Chen, Zhengfeng Lai, Haiming Gang, Kai Kang, and Afshin Dehghan.
\newblock Slowfast-llava: A strong training-free baseline for video large language models.
\newblock {\em arXiv preprint arXiv:2407.15841}, 2024.

\bibitem{videochat2}
Kunchang Li, Yali Wang, Yinan He, Yizhuo Li, Yi~Wang, Yi~Liu, Zun Wang, Jilan Xu, Guo Chen, Ping Luo, et~al.
\newblock Mvbench: A comprehensive multi-modal video understanding benchmark.
\newblock In {\em CVPR}, pages 22195--22206, 2024.

\bibitem{llamavid}
Yanwei Li, Chengyao Wang, and Jiaya Jia.
\newblock Llama-vid: An image is worth 2 tokens in large language models.
\newblock {\em arXiv preprint arXiv:2311.17043}, 2023.

\bibitem{videollama2}
Zesen Cheng, Sicong Leng, Hang Zhang, Yifei Xin, Xin Li, Guanzheng Chen, Yongxin Zhu, Wenqi Zhang, Ziyang Luo, Deli Zhao, et~al.
\newblock Videollama 2: Advancing spatial-temporal modeling and audio understanding in video-llms.
\newblock {\em arXiv preprint arXiv:2406.07476}, 2024.

\bibitem{videollama3}
Boqiang Zhang, Kehan Li, Zesen Cheng, Zhiqiang Hu, Yuqian Yuan, Guanzheng Chen, Sicong Leng, Yuming Jiang, Hang Zhang, Xin Li, et~al.
\newblock Videollama 3: Frontier multimodal foundation models for image and video understanding.
\newblock {\em arXiv preprint arXiv:2501.13106}, 2025.

\bibitem{timechat}
Shuhuai Ren, Linli Yao, Shicheng Li, Xu~Sun, and Lu~Hou.
\newblock Timechat: A time-sensitive multimodal large language model for long video understanding.
\newblock In {\em CVPR}, pages 14313--14323, 2024.

\bibitem{longvila}
Yukang Chen, Fuzhao Xue, Dacheng Li, Qinghao Hu, Ligeng Zhu, Xiuyu Li, Yunhao Fang, Haotian Tang, Shang Yang, Zhijian Liu, et~al.
\newblock Longvila: Scaling long-context visual language models for long videos.
\newblock {\em arXiv preprint arXiv:2408.10188}, 2024.

\bibitem{llavanextvid}
Yuanhan Zhang, Jinming Wu, Wei Li, Bo~Li, Zejun Ma, Ziwei Liu, and Chunyuan Li.
\newblock Video instruction tuning with synthetic data.
\newblock {\em arXiv preprint arXiv:2410.02713}, 2024.

\bibitem{videollava}
Bin Lin, Bin Zhu, Yang Ye, Munan Ning, Peng Jin, and Li~Yuan.
\newblock Video-llava: Learning united visual representation by alignment before projection.
\newblock {\em arXiv preprint arXiv:2311.10122}, 2023.

\bibitem{gemini15}
Machel Reid, Nikolay Savinov, Denis Teplyashin, Dmitry Lepikhin, Timothy Lillicrap, Jean-baptiste Alayrac, Radu Soricut, Angeliki Lazaridou, Orhan Firat, Julian Schrittwieser, et~al.
\newblock Gemini 1.5: Unlocking multimodal understanding across millions of tokens of context.
\newblock {\em arXiv preprint arXiv:2403.05530}, 2024.

\bibitem{internvl}
Zhe Chen, Jiannan Wu, Wenhai Wang, Weijie Su, Guo Chen, Sen Xing, Muyan Zhong, Qinglong Zhang, Xizhou Zhu, Lewei Lu, et~al.
\newblock Internvl: Scaling up vision foundation models and aligning for generic visual-linguistic tasks.
\newblock In {\em CVPR}, pages 24185--24198, 2024.

\bibitem{internvl25}
Zhe Chen, Weiyun Wang, Yue Cao, Yangzhou Liu, Zhangwei Gao, Erfei Cui, Jinguo Zhu, Shenglong Ye, Hao Tian, Zhaoyang Liu, et~al.
\newblock Expanding performance boundaries of open-source multimodal models with model, data, and test-time scaling.
\newblock {\em arXiv preprint arXiv:2412.05271}, 2024.

\bibitem{qwenvl}
Jinze Bai, Shuai Bai, Shusheng Yang, Shijie Wang, Sinan Tan, Peng Wang, Junyang Lin, Chang Zhou, and Jingren Zhou.
\newblock Qwen-vl: A versatile vision-language model for understanding, localization.
\newblock {\em Text Reading, and Beyond}, 2, 2023.

\bibitem{qwenvl25}
Shuai Bai, Keqin Chen, Xuejing Liu, Jialin Wang, Wenbin Ge, Sibo Song, Kai Dang, Peng Wang, Shijie Wang, Jun Tang, et~al.
\newblock Qwen2. 5-vl technical report.
\newblock {\em arXiv preprint arXiv:2502.13923}, 2025.

\bibitem{tong2024eyes}
Shengbang Tong, Zhuang Liu, Yuexiang Zhai, Yi~Ma, Yann LeCun, and Saining Xie.
\newblock Eyes wide shut? exploring the visual shortcomings of multimodal llms.
\newblock In {\em CVPR}, pages 9568--9578, 2024.

\bibitem{tvbench}
Daniel Cores, Michael Dorkenwald, Manuel Mucientes, Cees~GM Snoek, and Yuki~M Asano.
\newblock Tvbench: Redesigning video-language evaluation.
\newblock {\em arXiv preprint arXiv:2410.07752}, 2024.

\bibitem{rahmanzadehgervi2024vision}
Pooyan Rahmanzadehgervi, Logan Bolton, Mohammad~Reza Taesiri, and Anh~Totti Nguyen.
\newblock Vision language models are blind.
\newblock In {\em ACCV}, pages 18--34, 2024.

\bibitem{videohallucer}
Yuxuan Wang, Yueqian Wang, Dongyan Zhao, Cihang Xie, and Zilong Zheng.
\newblock Videohallucer: Evaluating intrinsic and extrinsic hallucinations in large video-language models.
\newblock {\em arXiv preprint arXiv:2406.16338}, 2024.

\bibitem{tempcompass}
Yuanxin Liu, Shicheng Li, Yi~Liu, Yuxiang Wang, Shuhuai Ren, Lei Li, Sishuo Chen, Xu~Sun, and Lu~Hou.
\newblock Tempcompass: Do video llms really understand videos?
\newblock {\em arXiv preprint arXiv:2403.00476}, 2024.

\bibitem{longva}
Peiyuan Zhang, Kaichen Zhang, Bo~Li, Guangtao Zeng, Jingkang Yang, Yuanhan Zhang, Ziyue Wang, Haoran Tan, Chunyuan Li, and Ziwei Liu.
\newblock Long context transfer from language to vision.
\newblock {\em arXiv preprint arXiv:2406.16852}, 2024.

\bibitem{ge2024v2pe}
Junqi Ge, Ziyi Chen, Jintao Lin, Jinguo Zhu, Xihui Liu, Jifeng Dai, and Xizhou Zhu.
\newblock V2pe: Improving multimodal long-context capability of vision-language models with variable visual position encoding.
\newblock {\em arXiv preprint arXiv:2412.09616}, 2024.

\bibitem{wei2024visual}
Hongchen Wei and Zhenzhong Chen.
\newblock Visual context window extension: A new perspective for long video understanding.
\newblock {\em arXiv preprint arXiv:2409.20018}, 2024.

\bibitem{mlvu}
Junjie Zhou, Yan Shu, Bo~Zhao, Boya Wu, Shitao Xiao, Xi~Yang, Yongping Xiong, Bo~Zhang, Tiejun Huang, and Zheng Liu.
\newblock Mlvu: A comprehensive benchmark for multi-task long video understanding.
\newblock {\em arXiv preprint arXiv:2406.04264}, 2024.

\bibitem{longvideobench}
Haoning Wu, Dongxu Li, Bei Chen, and Junnan Li.
\newblock Longvideobench: A benchmark for long-context interleaved video-language understanding.
\newblock {\em NeurIPS}, 37:28828--28857, 2025.

\bibitem{shu2025large}
Dong Shu, Haiyan Zhao, Jingyu Hu, Weiru Liu, Ali Payani, Lu~Cheng, and Mengnan Du.
\newblock Large vision-language model alignment and misalignment: A survey through the lens of explainability.
\newblock {\em arXiv preprint arXiv:2501.01346}, 2025.

\bibitem{dpa}
Pritam Sarkar, Sayna Ebrahimi, Ali Etemad, Ahmad Beirami, Sercan~{\"O} Ar{\i}k, and Tomas Pfister.
\newblock Data-augmented phrase-level alignment for mitigating object hallucination.
\newblock {\em arXiv preprint arXiv:2405.18654}, 2024.

\bibitem{zhou2023analyzing}
Yiyang Zhou, Chenhang Cui, Jaehong Yoon, Linjun Zhang, Zhun Deng, Chelsea Finn, Mohit Bansal, and Huaxiu Yao.
\newblock Analyzing and mitigating object hallucination in large vision-language models.
\newblock {\em arXiv preprint arXiv:2310.00754}, 2023.

\bibitem{alignment_lvlms}
Yiyang Zhou, Chenhang Cui, Rafael Rafailov, Chelsea Finn, and Huaxiu Yao.
\newblock Aligning modalities in vision large language models via preference fine-tuning.
\newblock {\em arXiv preprint arXiv:2402.11411}, 2024.

\bibitem{sun2023principle}
Zhiqing Sun, Yikang Shen, Qinhong Zhou, Hongxin Zhang, Zhenfang Chen, David Cox, Yiming Yang, and Chuang Gan.
\newblock Principle-driven self-alignment of language models from scratch with minimal human supervision.
\newblock {\em NeurIPS}, 36:2511--2565, 2023.

\bibitem{dpo}
Rafael Rafailov, Archit Sharma, Eric Mitchell, Christopher~D Manning, Stefano Ermon, and Chelsea Finn.
\newblock Direct preference optimization: Your language model is secretly a reward model.
\newblock {\em NeurIPS}, 36:53728--53741, 2023.

\bibitem{dpo_issues}
Yuzi Yan, Yibo Miao, Jialian Li, Yipin Zhang, Jian Xie, Zhijie Deng, and Dong Yan.
\newblock 3d-properties: Identifying challenges in dpo and charting a path forward.
\newblock {\em arXiv preprint arXiv:2406.07327}, 2024.

\bibitem{wang2024comprehensive}
Zhichao Wang, Bin Bi, Shiva~Kumar Pentyala, Kiran Ramnath, Sougata Chaudhuri, Shubham Mehrotra, Xiang-Bo Mao, Sitaram Asur, et~al.
\newblock A comprehensive survey of llm alignment techniques: Rlhf, rlaif, ppo, dpo and more.
\newblock {\em arXiv preprint arXiv:2407.16216}, 2024.

\bibitem{tdpo}
Yongcheng Zeng, Guoqing Liu, Weiyu Ma, Ning Yang, Haifeng Zhang, and Jun Wang.
\newblock Token-level direct preference optimization.
\newblock {\em arXiv preprint arXiv:2404.11999}, 2024.

\bibitem{schulman2015trust}
John Schulman, Sergey Levine, Pieter Abbeel, Michael Jordan, and Philipp Moritz.
\newblock Trust region policy optimization.
\newblock In {\em ICML}, pages 1889--1897. PMLR, 2015.

\bibitem{blip2}
Junnan Li, Dongxu Li, Silvio Savarese, and Steven Hoi.
\newblock Blip-2: Bootstrapping language-image pre-training with frozen image encoders and large language models.
\newblock {\em arXiv preprint arXiv:2301.12597}, 2023.

\bibitem{uniformerv2}
Kunchang Li, Yali Wang, Yinan He, Yizhuo Li, Yi~Wang, Limin Wang, and Yu~Qiao.
\newblock Uniformerv2: Spatiotemporal learning by arming image vits with video uniformer.
\newblock {\em arXiv preprint arXiv:2211.09552}, 2022.

\bibitem{ssv2}
Raghav Goyal, Samira Ebrahimi~Kahou, Vincent Michalski, Joanna Materzynska, Susanne Westphal, Heuna Kim, Valentin Haenel, Ingo Fruend, Peter Yianilos, Moritz Mueller-Freitag, et~al.
\newblock The" something something" video database for learning and evaluating visual common sense.
\newblock In {\em ICCV}, pages 5842--5850, 2017.

\bibitem{videochat}
KunChang Li, Yinan He, Yi~Wang, Yizhuo Li, Wenhai Wang, Ping Luo, Yali Wang, Limin Wang, and Yu~Qiao.
\newblock Videochat: Chat-centric video understanding.
\newblock {\em arXiv preprint arXiv:2305.06355}, 2023.

\bibitem{videochatgpt}
Muhammad Maaz, Hanoona Rasheed, Salman Khan, and Fahad~Shahbaz Khan.
\newblock Video-chatgpt: Towards detailed video understanding via large vision and language models.
\newblock {\em arXiv preprint arXiv:2306.05424}, 2023.

\bibitem{clevrr}
Justin Johnson, Bharath Hariharan, Laurens Van Der~Maaten, Li~Fei-Fei, C~Lawrence~Zitnick, and Ross Girshick.
\newblock Clevr: A diagnostic dataset for compositional language and elementary visual reasoning.
\newblock In {\em CVPR}, pages 2901--2910, 2017.

\bibitem{nextqa}
Junbin Xiao, Xindi Shang, Angela Yao, and Tat-Seng Chua.
\newblock Next-qa: Next phase of question-answering to explaining temporal actions.
\newblock In {\em CVPR}, pages 9777--9786, 2021.

\bibitem{ego4d}
Kristen Grauman, Andrew Westbury, Eugene Byrne, Zachary Chavis, Antonino Furnari, Rohit Girdhar, Jackson Hamburger, Hao Jiang, Miao Liu, Xingyu Liu, et~al.
\newblock Ego4d: Around the world in 3,000 hours of egocentric video.
\newblock In {\em CVPR}, pages 18995--19012, 2022.

\bibitem{tgif}
Yunseok Jang, Yale Song, Youngjae Yu, Youngjin Kim, and Gunhee Kim.
\newblock Tgif-qa: Toward spatio-temporal reasoning in visual question answering.
\newblock In {\em CVPR}, pages 2758--2766, 2017.

\bibitem{gpt4o}
Aaron Hurst, Adam Lerer, Adam~P Goucher, Adam Perelman, Aditya Ramesh, Aidan Clark, AJ~Ostrow, Akila Welihinda, Alan Hayes, Alec Radford, et~al.
\newblock Gpt-4o system card.
\newblock {\em arXiv preprint arXiv:2410.21276}, 2024.

\bibitem{bubeck2023sparks}
S{\'e}bastien Bubeck, Varun Chadrasekaran, Ronen Eldan, Johannes Gehrke, Eric Horvitz, Ece Kamar, Peter Lee, Yin~Tat Lee, Yuanzhi Li, Scott Lundberg, et~al.
\newblock Sparks of artificial general intelligence: Early experiments with gpt-4, 2023.

\bibitem{peng2023instruction}
Baolin Peng, Chunyuan Li, Pengcheng He, Michel Galley, and Jianfeng Gao.
\newblock Instruction tuning with gpt-4.
\newblock {\em arXiv preprint arXiv:2304.03277}, 2023.

\bibitem{zheng2023judging}
Lianmin Zheng, Wei-Lin Chiang, Ying Sheng, Siyuan Zhuang, Zhanghao Wu, Yonghao Zhuang, Zi~Lin, Zhuohan Li, Dacheng Li, Eric Xing, et~al.
\newblock Judging llm-as-a-judge with mt-bench and chatbot arena.
\newblock {\em NeurIPS}, 36:46595--46623, 2023.

\bibitem{chiang2023can}
Cheng-Han Chiang and Hung-yi Lee.
\newblock Can large language models be an alternative to human evaluations?
\newblock {\em arXiv preprint arXiv:2305.01937}, 2023.

\bibitem{lora}
Edward~J Hu, Yelong Shen, Phillip Wallis, Zeyuan Allen-Zhu, Yuanzhi Li, Shean Wang, Lu~Wang, Weizhu Chen, et~al.
\newblock Lora: Low-rank adaptation of large language models.
\newblock {\em ICLR}, 1(2):3, 2022.

\bibitem{vidhalluc}
Chaoyu Li, Eun~Woo Im, and Pooyan Fazli.
\newblock Vidhalluc: Evaluating temporal hallucinations in multimodal large language models for video understanding.
\newblock {\em arXiv preprint arXiv:2412.03735}, 2024.

\bibitem{videomme}
Chaoyou Fu, Yuhan Dai, Yongdong Luo, Lei Li, Shuhuai Ren, Renrui Zhang, Zihan Wang, Chenyu Zhou, Yunhang Shen, Mengdan Zhang, et~al.
\newblock Video-mme: The first-ever comprehensive evaluation benchmark of multi-modal llms in video analysis.
\newblock {\em arXiv preprint arXiv:2405.21075}, 2024.

\bibitem{rlhfv}
Tianyu Yu, Yuan Yao, Haoye Zhang, Taiwen He, Yifeng Han, Ganqu Cui, Jinyi Hu, Zhiyuan Liu, Hai-Tao Zheng, Maosong Sun, et~al.
\newblock Rlhf-v: Towards trustworthy mllms via behavior alignment from fine-grained correctional human feedback.
\newblock In {\em CVPR}, pages 13807--13816, 2024.

\bibitem{llava_tpo}
Rui Li, Xiaohan Wang, Yuhui Zhang, Zeyu Wang, and Serena Yeung-Levy.
\newblock Temporal preference optimization for long-form video understanding.
\newblock {\em arXiv preprint arXiv:2501.13919}, 2025.

\bibitem{minicpm_v}
Yuan Yao, Tianyu Yu, Ao~Zhang, Chongyi Wang, Junbo Cui, Hongji Zhu, Tianchi Cai, Haoyu Li, Weilin Zhao, Zhihui He, et~al.
\newblock Minicpm-v: A gpt-4v level mllm on your phone.
\newblock {\em arXiv preprint arXiv:2408.01800}, 2024.

\bibitem{nvila}
Zhijian Liu, Ligeng Zhu, Baifeng Shi, Zhuoyang Zhang, Yuming Lou, Shang Yang, Haocheng Xi, Shiyi Cao, Yuxian Gu, Dacheng Li, et~al.
\newblock Nvila: Efficient frontier visual language models.
\newblock {\em arXiv preprint arXiv:2412.04468}, 2024.

\bibitem{qwen2vl}
Peng Wang, Shuai Bai, Sinan Tan, Shijie Wang, Zhihao Fan, Jinze Bai, Keqin Chen, Xuejing Liu, Jialin Wang, Wenbin Ge, et~al.
\newblock Qwen2-vl: Enhancing vision-language model's perception of the world at any resolution.
\newblock {\em arXiv preprint arXiv:2409.12191}, 2024.

\bibitem{zhang2024llava}
Y~Zhang, B~Li, H~Liu, Y~Lee, L~Gui, D~Fu, J~Feng, Z~Liu, and C~Li.
\newblock Llava-next: A strong zero-shot video understanding model.
\newblock 2024.

\bibitem{sharegpt4video}
Lin Chen, Xilin Wei, Jinsong Li, Xiaoyi Dong, Pan Zhang, Yuhang Zang, Zehui Chen, Haodong Duan, Zhenyu Tang, Li~Yuan, et~al.
\newblock Sharegpt4video: Improving video understanding and generation with better captions.
\newblock {\em NeurIPS}, 37:19472--19495, 2024.

\bibitem{pllava}
Lin Xu, Yilin Zhao, Daquan Zhou, Zhijie Lin, See~Kiong Ng, and Jiashi Feng.
\newblock Pllava: Parameter-free llava extension from images to videos for video dense captioning.
\newblock {\em arXiv preprint arXiv:2404.16994}, 2024.

\bibitem{aria}
Dongxu Li, Yudong Liu, Haoning Wu, Yue Wang, Zhiqi Shen, Bowen Qu, Xinyao Niu, Fan Zhou, Chengen Huang, Yanpeng Li, et~al.
\newblock Aria: An open multimodal native mixture-of-experts model.
\newblock {\em arXiv preprint arXiv:2410.05993}, 2024.

\bibitem{gpt4}
Josh Achiam, Steven Adler, Sandhini Agarwal, Lama Ahmad, Ilge Akkaya, Florencia~Leoni Aleman, Diogo Almeida, Janko Altenschmidt, Sam Altman, Shyamal Anadkat, et~al.
\newblock Gpt-4 technical report.
\newblock {\em arXiv preprint arXiv:2303.08774}, 2023.

\bibitem{umt}
Kunchang Li, Yali Wang, Yizhuo Li, Yi~Wang, Yinan He, Limin Wang, and Yu~Qiao.
\newblock Unmasked teacher: Towards training-efficient video foundation models.
\newblock In {\em ICCV}, pages 19948--19960, 2023.

\bibitem{dinov2}
Maxime Oquab, Timoth{\'e}e Darcet, Th{\'e}o Moutakanni, Huy Vo, Marc Szafraniec, Vasil Khalidov, Pierre Fernandez, Daniel Haziza, Francisco Massa, Alaaeldin El-Nouby, et~al.
\newblock Dinov2: Learning robust visual features without supervision.
\newblock {\em arXiv preprint arXiv:2304.07193}, 2023.

\bibitem{siglip}
Xiaohua Zhai, Basil Mustafa, Alexander Kolesnikov, and Lucas Beyer.
\newblock Sigmoid loss for language image pre-training.
\newblock In {\em ICCV}, pages 11975--11986, 2023.

\bibitem{video_star}
Orr Zohar, Xiaohan Wang, Yonatan Bitton, Idan Szpektor, and Serena Yeung-Levy.
\newblock Video-star: Self-training enables video instruction tuning with any supervision.
\newblock {\em arXiv preprint arXiv:2407.06189}, 2024.

\bibitem{webvid}
Max Bain, Arsha Nagrani, G{\"u}l Varol, and Andrew Zisserman.
\newblock Frozen in time: A joint video and image encoder for end-to-end retrieval.
\newblock In {\em ICCV}, pages 1728--1738, 2021.

\bibitem{moviechat}
Enxin Song, Wenhao Chai, Guanhong Wang, Yucheng Zhang, Haoyang Zhou, Feiyang Wu, Haozhe Chi, Xun Guo, Tian Ye, Yanting Zhang, et~al.
\newblock Moviechat: From dense token to sparse memory for long video understanding.
\newblock In {\em CVPR}, pages 18221--18232, 2024.

\bibitem{videochat_flash}
Xinhao Li, Yi~Wang, Jiashuo Yu, Xiangyu Zeng, Yuhan Zhu, Haian Huang, Jianfei Gao, Kunchang Li, Yinan He, Chenting Wang, et~al.
\newblock Videochat-flash: Hierarchical compression for long-context video modeling.
\newblock {\em arXiv preprint arXiv:2501.00574}, 2024.

\bibitem{lita}
De-An Huang, Shijia Liao, Subhashree Radhakrishnan, Hongxu Yin, Pavlo Molchanov, Zhiding Yu, and Jan Kautz.
\newblock Lita: Language instructed temporal-localization assistant.
\newblock In {\em ECCV}, pages 202--218. Springer, 2024.

\bibitem{videogptplus}
Muhammad Maaz, Hanoona Rasheed, Salman Khan, and Fahad Khan.
\newblock Videogpt+: Integrating image and video encoders for enhanced video understanding.
\newblock {\em arXiv preprint arXiv:2406.09418}, 2024.

\bibitem{internvideo2}
Yi~Wang, Kunchang Li, Xinhao Li, Jiashuo Yu, Yinan He, Guo Chen, Baoqi Pei, Rongkun Zheng, Jilan Xu, Zun Wang, et~al.
\newblock Internvideo2: Scaling video foundation models for multimodal video understanding.
\newblock {\em arXiv preprint arXiv:2403.15377}, 2024.

\bibitem{clip}
Alec Radford, Jong~Wook Kim, Chris Hallacy, Aditya Ramesh, Gabriel Goh, Sandhini Agarwal, Girish Sastry, Amanda Askell, Pamela Mishkin, Jack Clark, et~al.
\newblock Learning transferable visual models from natural language supervision.
\newblock In {\em ICML}, pages 8748--8763. PmLR, 2021.

\bibitem{rlhf}
Paul~F Christiano, Jan Leike, Tom Brown, Miljan Martic, Shane Legg, and Dario Amodei.
\newblock Deep reinforcement learning from human preferences.
\newblock {\em NeurIPS}, 30, 2017.

\bibitem{ppo}
John Schulman, Filip Wolski, Prafulla Dhariwal, Alec Radford, and Oleg Klimov.
\newblock Proximal policy optimization algorithms.
\newblock {\em arXiv preprint arXiv:1707.06347}, 2017.

\bibitem{video_llm_survey2}
Neelu Madan, Andreas M{\o}gelmose, Rajat Modi, Yogesh~S Rawat, and Thomas~B Moeslund.
\newblock Foundation models for video understanding: A survey.
\newblock {\em arXiv preprint arXiv:2405.03770}, 2024.

\bibitem{video_llm_survey}
Yunlong Tang, Jing Bi, Siting Xu, Luchuan Song, Susan Liang, Teng Wang, Daoan Zhang, Jie An, Jingyang Lin, Rongyi Zhu, et~al.
\newblock Video understanding with large language models: A survey.
\newblock {\em arXiv preprint arXiv:2312.17432}, 2023.

\bibitem{llava}
Haotian Liu, Chunyuan Li, Qingyang Wu, and Yong~Jae Lee.
\newblock Visual instruction tuning.
\newblock {\em NeurIPS}, 36:34892--34916, 2023.

\bibitem{instructblip}
Wenliang Dai, Junnan Li, Dongxu Li, Anthony Meng~Huat Tiong, Junqi Zhao, Weisheng Wang, Boyang Li, Pascale Fung, and Steven Hoi.
\newblock Instructblip: Towards general-purpose vision-language models with instruction tuning.
\newblock {\em arXiv preprint arXiv:2305.06500}, 2023.

\bibitem{qwen25vl}
Shuai Bai, Keqin Chen, Xuejing Liu, Jialin Wang, Wenbin Ge, Sibo Song, Kai Dang, Peng Wang, Shijie Wang, Jun Tang, et~al.
\newblock Qwen2. 5-vl technical report.
\newblock {\em arXiv preprint arXiv:2502.13923}, 2025.

\bibitem{zhang2025mm}
Yi-Fan Zhang, Tao Yu, Haochen Tian, Chaoyou Fu, Peiyan Li, Jianshu Zeng, Wulin Xie, Yang Shi, Huanyu Zhang, Junkang Wu, et~al.
\newblock Mm-rlhf: The next step forward in multimodal llm alignment.
\newblock {\em arXiv preprint arXiv:2502.10391}, 2025.

\bibitem{videosavi}
Yogesh Kulkarni and Pooyan Fazli.
\newblock Videosavi: Self-aligned video language models without human supervision.
\newblock {\em arXiv preprint arXiv:2412.00624}, 2024.

\bibitem{isrt}
Daechul Ahn, Yura Choi, San Kim, Youngjae Yu, Dongyeop Kang, and Jonghyun Choi.
\newblock i-srt: Aligning large multimodal models for videos by iterative self-retrospective judgment.
\newblock {\em arXiv preprint arXiv:2406.11280}, 2024.

\bibitem{zero_offload}
Jie Ren, Samyam Rajbhandari, Reza~Yazdani Aminabadi, Olatunji Ruwase, Shuangyan Yang, Minjia Zhang, Dong Li, and Yuxiong He.
\newblock Zero-offload: Democratizing billion-scale model training.
\newblock In {\em USENIX ATC}, pages 551--564, 2021.

\bibitem{zero_offload2}
Samyam Rajbhandari, Olatunji Ruwase, Jeff Rasley, Shaden Smith, and Yuxiong He.
\newblock Zero-infinity: Breaking the gpu memory wall for extreme scale deep learning.
\newblock In {\em SC}, pages 1--14, 2021.

\end{thebibliography}
}

\clearpage
\appendix

\begin{center}
    \maketitle
    \Large
    \textbf{Appendix}
\end{center}

\setcounter{table}{0}
\setcounter{figure}{0}
\setcounter{equation}{0}
\renewcommand{\theequation}{S\arabic{equation}}
\renewcommand{\thetable}{S\arabic{table}}
\renewcommand\thefigure{S\arabic{figure}}

\section{RRPO Pseudocode (PyTorch Style)} \label{supsec:pseudocode}

\begin{verbatim}
import torch
import torch.nn.functional as F

def rrpo_loss(self, logits, ref_logits, phrase_ids, alpha, beta):
    """
    logits:         logits from pi_theta. 
                    shape: (batch_size, sequence_length, vocab_size)
    ref_logits:     logits from pi_ref. 
                    shape: (batch_size, sequence_length, vocab_size)
    phrase_ids:     phrase identifiers where tokens belonging to the same
                    phrase share the same value; additionally, they remain 
                    the same between correct and misaligned phrases to 
                    maintain correspondence.   
                    shape: (batch_size, sequence_length)
    correct_idx:    indices of the correct responses in batch
    wrong_idx:      indices of the wrong responses in batch
    alpha:          coefficient to control the token-wise KL divergence
    beta:           coefficient to control reward/penalty
    """

    # compute log probabilities
    logps = logits.log_softmax(dim=-1)
    ref_ps = ref_logits.softmax(dim=-1)
    ref_logps = ref_ps.log()

    # compute token-wise KL divergence
    token_wise_kl = (ref_ps * (ref_logps - logps)).sum(dim=-1)
    
    # compute the margin
    logps_margin = logps - ref_logps
    
    # accumulate log probabilities of the phrases with key concepts
    # here 0 indicates ignored tokens
    unique_phrase_ids = torch.unique(phrase_ids, sorted=True)[1:]
    phrase_logps_margin = torch.zeros(
            phrase_ids.size(0), len(unique_phrase_ids))
    for i, phrase_id in enumerate(unique_phrase_ids):
        mask = (phrase_ids == phrase_id).float()
        phrase_logps_margin[:, i] = (logps_margin * mask).sum(dim=-1)
           
    chosen_logps_margin = phrase_logps_margin[correct_idx]
    rejected_logps_margin = phrase_logps_margin[wrong_idx]
    logits_margin = (chosen_logps_margin -          
                        rejected_logps_margin).sum(dim=-1)

    chosen_token_wise_kl = token_wise_kl.sum(dim=-1)[chosen_idx]
    
    losses = -torch.logsigmoid(beta * logits_margin) 
             + alpha * chosen_token_wise_kl
    
    return losses
\end{verbatim}
\clearpage

\begin{figure}%
    \centering
    \includegraphics[width=0.5\linewidth]{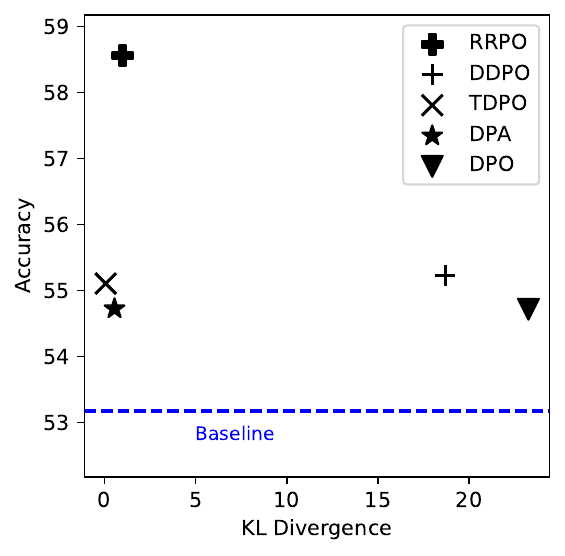}
    \captionof{figure}{Comparison of characteristics among preference optimization methods. Both DDPO and DPO diverge significantly from their initial state after alignment. While TDPO and DPA are effective in restricting model divergence, they are less effective in improving performance. RRPO achieves excellent performance with minimal divergence from the base model.
    }
    \label{fig:score_vs_kld_redo}
\end{figure}

\section{Comparing Loss Formulation of RRPO and Other Methods} \label{supsec:ablation_variants_details}
This section presents a comparative analysis between RRPO and other preference optimization methods 
studied in this work, i.e., DPO, DDPO, TDPO, and DPA. 
We begin by outlining the respective loss functions, followed by a detailed discussion of their similarities and differences.

\noindent\textbf{Comparison with DPO.}

As previously discussed, the DPO \cite{dpo} loss function is defined as:
\begin{equation}
\begin{aligned}
\mathcal{L}_\text{DPO}(\pi_{\theta};\pi_\text{ref}) 
& = -\mathbb{E} 
\biggl[ \log \sigma 
\bigl(r_\theta(x,y^+)-r_\theta(x,y^-)\bigr) 
\biggr], \\
& = -\mathbb{E} 
\biggl[ \log \sigma 
\Bigl(
\beta \log \frac{\pi_\theta(y^+|x)}{\pi_\text{ref}(y^+|x)}
- \beta \log \frac{\pi_\theta(y^-|x)}{\pi_\text{ref}(y^-|x)}
\Bigr)
\biggr].
\end{aligned}
\end{equation}
The DPO loss function calculates reward over all tokens in $y^+$ and $y^-$, despite the fact that there might be a few sub-sequences that are conceptually different. This approach results in coarse-grained reward modeling, and because it penalizes all tokens in the response, the loss function accumulates a large gradient and tends to diverge significantly from the base model, potentially resulting in weak alignment, as shown in Figure \ref{fig:score_vs_kld_redo}. RRPO is introduced to address two key challenges of DPO: to provide fine-grained feedback and restrict the divergence of the model away from its initial state.

\noindent\textbf{Comparison with DDPO.}

{DDPO} \cite{rlhfv} extends DPO by incorporating a weighted reward based on the sub-sequence level differences between $y^+$ and $y^-$, and is defined as:
\begin{equation}
\begin{aligned}
\mathcal{L}_\text{DDPO}(\pi_{\theta};\pi_\text{ref}) 
& = -\mathbb{E} 
\biggl[ \log \sigma 
\Bigl(
\beta \log \frac{\pi_\theta(y^+|x)}{\pi_\text{ref}(y^+|x)}
- \beta \log \frac{\pi_\theta(y^-|x)}{\pi_\text{ref}(y^-|x)}
\Bigr)
\biggr],
\end{aligned}
\end{equation}
where $\log \pi(y|x) = \sum \limits_{y_i\in y} \log p(y_i|x, y_{<i})$ is modified as: 
\begin{equation*}
\log \pi(y|x) = \frac{1}{N} \bigl[\sum \limits_{y_i\in y_\text{same}} \log p(y_i|x, y_{<i}) + \gamma \sum \limits_{y_i \in y_\text{different}} \log p(y_i|x, y_{<i})\bigr].
\end{equation*}
Here, $y_\text{same}$ and $y_\text{different}$ indicate unchanged and changed segments between $y^+$ and $y^-$. Moreover, $\gamma > 1$ is a weighting hyperparameter, and larger $\gamma$ indicates more weight on those changed segments.
While DDPO is designed to provide fine-grained feedback, its loss formulation is not as effective as RRPO and is also prone to diverging far from its initial state, similar to DPO, due to weak regularization, see Figure \ref{fig:score_vs_kld_redo}.

\textbf{Comparison with TDPO.}

{TDPO} \cite{tdpo} is also derived from DPO, incorporating an additional regularization term ($\mathbb{D}_\text{TKL}$) between $\pi_\theta$ and $\pi_\text{ref}$, which is defined as:
\begin{equation}
\begin{aligned}
\mathcal{L}_\text{TDPO}(\pi_{\theta};\pi_\text{ref}) = &
-\mathbb{E} 
\biggl[ \log \sigma 
\biggl(
\bigl(r_\theta(x,y^+)-r_\theta(x,y^-)\bigr) 
\\
& 
- \alpha
\Bigl(
\beta \mathbb{D}_{\mathrm{TKL}}\left({x},{y}_w;\pi_{\mathrm{ref}}\| \pi_{\theta}\right)
-\mathnormal{sg} \bigl(\beta \mathbb{D}_{\mathrm{TKL}}\left({x},{y}_c;\pi_{\mathrm{ref}}\| \pi_{\theta}\right)\bigr)
\Bigr)
\biggr)
\biggr].
\end{aligned}
\end{equation}
As shown in Figure \ref{fig:score_vs_kld_redo}, TDPO is effective in restricting the divergence of the base model, but its performance is almost the same as DPO and falls short of our RRPO.

\textbf{Comparison with DPA.} 

DPA \cite{dpa} is a phrase-level alignment method unlike DPO and its variants. The DPA loss is composed of two terms where the first term computes the relative log-probability between two phrases of $y^+$ and $y^-$, and the second term works as a regularizer between $\pi_\theta$ and $\pi_\text{ref}$, formulated as:
\begin{equation}
\begin{aligned}
\mathcal{L}_\text{DPA}(\pi_{\theta};\pi_\text{ref}) %
=&-\mathbb{E}
\biggl[
\frac{1}{N}\sum_{i=1}^{N}
- \log\frac
{
P(y^+_i)
}
{
P(y^+_i)+P(y^-_i)
}
+ 
\alpha\!\cdot\!\mathbb{D}_{\text{TKL}} \big(x,y^+;\pi_{\mathrm{ref}}\| \pi_{\theta}\big)
\biggr],
\end{aligned}
\end{equation}
where $P(y^+_i)$ and $P(y^-_i)$ denote the probability of $i$-th phrase in $y^+$ and $y^-$. Assume, $y$ is expressed as a sequence of tokens $\{ t_1, t_2, \ldots, t_{|T|} \}$, then the probability of $i$-th phrase can be computed as
$
\prod\limits_{j=s_i}^{e_i} 
\pi_\theta(t_j|x, t_{<j})
$, where $s_i$ and $e_i$ represent the start and end token indices.

The loss formulation of RRPO draws inspiration from DPA to achieve fine-grained alignment without the risk of divergence. However, we identify a fundamental limitation in DPA: it directly adjusts the probabilities of $\pi_\theta$ to modify the probability ratio between preferred and non-preferred phrases. This approach leads to an inaccurate probability ratio after the initial pair of preferred and non-preferred segments, as subsequent segment probabilities become dependent on their preceding elements. Therefore, DPA is not accurate in providing fine-grained feedback for
sequences composed of multiple sub-sequences of key concepts. As shown in Figure \ref{fig:score_vs_kld_redo}, DPA, while successful in controlling divergence, is ineffective in improving performance.

\clearpage
\section{Additional Results} 
\label{supsec:addl_results}

\textbf{Statistical analysis between RRPO and DPO.}
\label{supsec:stat_anal}
We perform statistical analysis between RRPO and DPO aligned model. We consider the performance (denoted as \texttt{Score}) difference, whether an improvement or decline, of the \texttt{Model1} relative to the \texttt{Model2} on a specific task to be statistically significant if the Adjusted $\Delta$ exceeds zero, where $\Delta$ denotes the difference in performance between the \texttt{Model1} and \texttt{Model2}. Statistical significance is assessed using the Standard Error (SE) at a $95\%$ confidence level. The corresponding mathematical formulations are presented below.
\begin{equation*}
\begin{split}
\texttt{SE}=\sqrt{\frac{\texttt{Score}_\texttt{Model1} \times (1 - \texttt{Score}_\texttt{Model2})}{\text{number of samples}}},\\
\Delta=\texttt{Score}_\texttt{Model1}-\texttt{Score}_\texttt{Model2},\\
\text{Adjusted }\Delta=\Delta-1.96*\text{SE}.
\end{split}
\end{equation*}

The results presented in Table \ref{tab:stat_anal} shows that RRPO almost always outperforms DPO and even in several cases improvements are statistically significant.

\begin{table}[!h]%
\centering
    \fontsize{9pt}{10pt}\selectfont
    \setlength\tabcolsep{4pt}
    \caption{Statistical analysis between RRPO and DPO variants. \green{Green underline} indicates that the performance improvements are statistically significant. Also, there are no instances in which the DPO variants achieve statistically significant gains over RRPO.
    }
    \label{tab:stat_anal}
    \resizebox{\linewidth}{!}{
    \begin{tabular}{lcccccccc}
    \toprule
    \bf Model
    & \rotatebox{0}{\specialcellleft{\bf TV\\\bf Bench}} 
    & \rotatebox{0}{\specialcellleft{\bf TempCo-\\\bf mpass$_\text{Avg}$}} 
    & \rotatebox{0}{\specialcellleft{\bf Video\\\bf Hallucer}} 
    & \rotatebox{0}{\specialcellleft{\bf Vid\\\bf Halluc}} 
    & \rotatebox{0}{\specialcellleft{\bf MV\\\bf Bench}} 
    & \rotatebox{0}{\specialcellleft{\bf Video\\\bf MME}} 
    & \rotatebox{0}{\specialcellleft{\bf MLVU$_\text{val}$\\ \bf \small (M-Avg)}} 
    & \rotatebox{0}{\specialcellleft{\bf LongVideo\\\bf Bench$_\text{Val}$}}  \\
    \midrule\midrule

VideoChat2$_\text{7B}$+DPO 	 & $45.7$ 	 & $60.0$ 	 & $22.1$ 	 & $72.4$ 	 & $59.7$ 	 & $43.0$ 	 & $47.4$ 	 & $41.0$ \\
VideoChat2$_\text{7B}$+RRPO 	 & $45.8$ 	 & $60.2$ 	 & $\greenuline{32.9}$ 	 & $\greenuline{76.4}$ 	 & $59.1$ 	 & $44.3$ 	 & $47.9$ 	 & $42.8$ \\ \midrule
LLaVA-Video$_\text{7B}$+DPO 	 & $51.9$ 	 & $66.4$ 	 & $53.3$ 	 & $76.5$ 	 & $60.2$ 	 & $63.1$ 	 & $67.4$ 	 & $59.4$ \\
LLaVA-Video$_\text{7B}$+RRPO 	 & $51.9$ 	 & $66.8$ 	 & $55.7$ 	 & $76.5$ 	 & $62.0$ 	 & $64.5$ 	 & $69.1$ 	 & $60.4$ \\ \midrule
LLaVA-Video$_\text{7B}$+TPO 	 & $51.1$ 	 & $66.1$ 	 & $50.6$ 	 & $76.3$ 	 & $60.6$ 	 & $65.6$ 	 & $68.9$ 	 & $60.1$ \\
LLaVA-Video$_\text{7B}$+RRPO 	 & $52.2$ 	 & $67.4$ 	 & $\greenuline{55.8}$ 	 & $76.6$ 	 & $62.0$ 	 & $65.5$ 	 & $69.4$ 	 & $61.3$ \\ \midrule
LongVU$_\text{7B}$+DPO 	 & $54.3$ 	 & $64.3$ 	 & $40.9$ 	 & $68.5$ 	 & $65.8$ 	 & $56.6$ 	 & $63.6$ 	 & $49.4$ \\
LongVU$_\text{7B}$+RRPO 	 & $\greenuline{56.5}$ 	 & $64.5$ 	 & $\greenuline{44.0}$ 	 & $\greenuline{71.7}$ 	 & $66.7$ 	 & $57.7$ 	 & $64.5$ 	 & $49.7$ \\

    \bottomrule
    \end{tabular}   
}
\end{table}

\clearpage
\textbf{Detailed results.}
This section details the results for the subcategories of the evaluation benchmarks used in our study.

\begin{table}[!h]
\fontsize{9pt}{10pt}\selectfont
\centering
\caption{Detailed results on TempCompass.}
\label{tab:detailed_tempcompass}
\begin{tabular}{lccccc}
\toprule
Models	 & Caption Matching	 & Captioning	 & Multi-choice	 & Yes-No	 & Avg. \\
\midrule\midrule
VideoChat2$_\text{7B}$  	 & $69.5 $  	 & $46.6 $  	 & $58.0 $  	 & $63.0 $  	 & $59.3 $ \\
~~~+ RRPO  	 & $73.2 $  	 & $48.5 $  	 & $56.6 $  	 & $62.6 $  	 & $60.2 $ \\ \midrule
LlavaVideo$_\text{7B}$  	 & $75.1 $  	 & $50.2 $  	 & $67.6 $  	 & $71.0 $  	 & $66.0 $ \\
~~~+ RRPO  	 & $75.8 $  	 & $52.0 $  	 & $68.6 $  	 & $70.9 $  	 & $66.8 $ \\
~~~+ RRPO ($32$f)  	 & $76.6 $  	 & $53.0 $  	 & $68.7 $  	 & $71.3 $  	 & $67.4 $ \\ \midrule

LongVU$_\text{7B}$  	 & $74.7 $  	 & $49.8 $  	 & $63.9 $  	 & $67.3 $  	 & $63.9 $ \\
~~~+ RRPO  	 & $75.2 $  	 & $50.6 $  	 & $64.7 $  	 & $67.4 $  	 & $64.5 $ \\ 

\bottomrule
\end{tabular}
\end{table}

\begin{table}[!h]
\fontsize{9pt}{10pt}\selectfont
\centering
\caption{Detailed results on VideoHallucer.}
\label{tab:detailed_videohallucer}
\begin{tabular}{lcccccc}
\toprule
Model	 & Object relation	 & Temporal	 & Semantic	 & Factual	 & Non-factual	 & Avg. \\
\midrule\midrule
VideoChat2$_\text{7B}$  	 & $47.5 $  	 & $8.0 $  	 & $38.5 $  	 & $1.0 $  	 & $20.5 $  	 & $23.1 $ \\
~~~+ RRPO  	 & $53.5 $  	 & $24.0 $  	 & $55.0 $  	 & $5.0 $  	 & $27.0 $  	 & $32.9 $ \\ \midrule
LlavaVideo$_\text{7B}$  	 & $66.0 $  	 & $56.5 $  	 & $65.5 $  	 & $13.5 $  	 & $48.5 $  	 & $50.0 $ \\
~~~+ RRPO  	 & $65.5 $  	 & $65.5 $  	 & $71.0 $  	 & $23.5 $  	 & $53.0 $  	 & $55.7 $ \\
~~~+ RRPO ($32$f)  	 & $65.5 $  	 & $65.5 $  	 & $71.5 $  	 & $23.5 $  	 & $53.0 $  	 & $55.8 $ \\
\midrule
LongVU$_\text{7B}$  	 & $50.5 $  	 & $46.0 $  	 & $43.0 $  	 & $17.0 $  	 & $39.5 $  	 & $39.2 $ \\ 
~~~+ RRPO  	 & $53.0 $  	 & $48.0 $  	 & $50.0 $  	 & $26.0 $  	 & $43.0 $  	 & $44.0 $ \\ 
\bottomrule
\end{tabular}
\end{table}

\begin{table}[!h]
\fontsize{9pt}{10pt}\selectfont
\centering
\caption{Detailed results on VidHalluc.}
\label{tab:detailed_vidhalluc}
\begin{tabular}{lcccc}
\toprule
Model	 & BinaryQA	 & MCQ	 &  Scene Transition & Avg. \\
\midrule\midrule
VideoChat2$_\text{7B}$  	 & $66.8 $  	 & $84.9 $  	 & $68.2 $  	 & $73.3 $ \\
~~~+ RRPO  	 & $72.7 $  	 & $85.5 $  	 & $70.9 $  	 & $76.4 $ \\ \midrule

LlavaVideo$_\text{7B}$   	 & $77.9 $  	 & $91.4 $  	 & $60.6 $  	 & $76.6 $ \\
~~~+ RRPO  	 & $78.4 $  	 & $91.6 $  	 & $59.5 $  	 & $76.5 $ \\
~~~+ RRPO ($32$f)  	 & $78.6 $  	 & $91.7 $  	 & $59.5 $  	 & $76.6 $ \\ \midrule
LongVU$_\text{7B}$  	 & $71.4 $  	 & $87.0 $  	 & $43.4 $  	 & $67.3 $ \\
~~~+ RRPO  	 & $74.2 $  	 & $88.2 $  	 & $52.7 $  	 & $71.7 $ \\ 

\bottomrule
\end{tabular}
\end{table}

\begin{table}[!h]
\fontsize{9pt}{10pt}\selectfont
\centering
\caption{Detailed results on VideoMME.}
\label{tab:detailed_vidmme}
\begin{tabular}{lcccc}
\toprule
Model	 & Short	 & Medium	 &  Long & Avg. \\
\midrule\midrule
VideoChat2$_\text{7B}$  	 & $49.0 $  	 & $38.6 $  	 & $35.6 $  	 & $41.0 $ \\
~~~+ RRPO  	 & $52.2 $  	 & $41.9 $  	 & $38.8 $  	 & $44.3 $ \\ \midrule
LlavaVideo$_\text{7B}$  	 & $76.3 $  	 & $62.8 $  	 & $52.8 $  	 & $64.0 $ \\
~~~+ RRPO  	 & $76.6 $  	 & $63.1 $  	 & $53.8 $  	 & $64.5 $ \\
~~~+ RRPO ($32$f)  	 & $76.7 $  	 & $62.9 $  	 & $53.9 $  	 & $64.5 $ \\ \midrule
LongVU$_\text{7B}$  	 & $66.1 $  	 & $54.7 $  	 & $47.9 $  	 & $56.2 $ \\
~~~+ RRPO  	 & $67.7 $  	 & $55.0 $  	 & $50.3 $  	 & $57.7 $ \\

\bottomrule
\end{tabular}
\end{table}

\clearpage
\section{Additional Details of Training Data} \label{supsec:details_data}

\textbf{Prompt templates.}

The instructions used in processing open-ended generated responses employing GPT4o are presented in Figures \ref{fig:prompt_stage1_open} and \ref{fig:prompt_stage2_open}.

\begin{sourcecode}{}
% {Prompt used in open-ended response processing stage 1}
Thoroughly read the question and the given answers.

Your task is to determine whether the "Predicted answer" is "Correct" or "Wrong" based on the "Question" and "Reference answer".
To determine correctness, focus on the key aspects in the answers, such as objects, actions, and their attributes, among others.
The "Predicted answer" may have partial information in comparison to the "Reference answer", in that case check whether at least the partial information can be fully verified based on the "Reference answer".

Please respond with any of the following and nothing else:

- "Correct" if the predicted answer is correct based on the reference answer.
- "Wrong" if the predicted answer is not fully correct based on the reference answer.
- "Undecided" if you are not sure about their correctness.

Question: {question}
Reference answer: {ground_truth}
Predicted answer: {generated_response}
\end{sourcecode}
\captionof{figure}{Prompt used in open-ended response processing stage 1.}
\label{fig:prompt_stage1_open}

\begin{sourcecode}{}
% {Prompt used in open-ended response processing stage 2}
***Turn 1***

Identify the key differences between these two sentences.
To identify differences focus on the key aspects in the sentences, 
such as objects, actions, and their attributes, among others.
If there are no key difference between these two sentences, 
please respond with "None" and nothing else.

Sentence 1: {sentence_from_ground_truth}
Sentence 2: {sentence_from_generated_response}
    
***Turn 2***

Please rewrite the "Sentence 1" by incorporating the differences you 
mentioned earlier. Your final response should contain only the revised 
sentence and nothing else.

Sentence 1: {sentence_from_ground_truth}
\end{sourcecode}
\captionof{figure}{Prompt used in open-ended response processing stage 2.}
\label{fig:prompt_stage2_open}



\clearpage
\section{Implementation Details} \label{supsec:implementation_details}

\noindent\textbf{Training hyperparameters.}

\begin{table}[!h]
\caption{Details of training hyperparameters.}
\label{tab:impl_details}
\centering
\resizebox{0.85\textwidth}{!}{
    \begin{tabular}{lccc}
    \toprule
         & \bf VideoChat2 & \bf LLaVA-Video & \bf LongVU \\
    \midrule\midrule
    LLM & Mistral & Qwen2 & Qwen2 \\
    Vision encoder  & UMT & SigLIP & SigLIP+DINOv2 \\
    Trainable module & \multicolumn{3}{c}{LoRA in LLM and everything else is kept frozen} \\
    LoRA setup \cite{lora} & \multicolumn{3}{c}{rank=128, alpha=256} \\
    Learning rate & 2e-5 & 5e-6 & 5e-6 \\
    Learning rate scheduler & Cosine & Cosine & Cosine \\
    Optimizer & AdamW & AdamW & AdamW \\
    Weight decay & 0.02 & 0.0 & 0.0 \\
    Warmup ratio & - & 0.03 & 0.03 \\
    Epoch & 1 & 1 & 1\\
    Batch size per GPU & 2 & 1 & 1 \\
    Batch size (total) & 32 & 32 & 32 \\
    $\alpha$ (loss coefficient) & 0.01 & 0.01 & 0.05 \\
    $\beta$ (loss coefficient) & 0.9 & 0.1 & 0.5 \\
    Memory optimization & - & Zero stage 3 \cite{zero_offload,zero_offload2} & FSDP \\
    \bottomrule
    \end{tabular}
}
\end{table}

\noindent\textbf{Licenses of existing assets used.}

\begin{itemize}[leftmargin=10pt] %
    \item VideoChat2 (Apache License 2.0): \textit{https://huggingface.co/OpenGVLab/VideoChat2\_stage3\_Mistral\_7B}
    \item LLaVA-Video (Apache License 2.0 ): \textit{https://huggingface.co/lmms-lab/LLaVA-Video-7B-Qwen2}
    \item LongVU (Apache License 2.0): \textit{https://huggingface.co/Vision-CAIR/LongVU\_Qwen2\_7B}
    \item VideoChat-IT (MIT): \textit{https://huggingface.co/datasets/OpenGVLab/VideoChat2-IT}
\end{itemize}

\section{Broader impact} \label{suppsec:disc}



As generative models are increasingly deployed in real-world applications, there is a growing need for post-training methods that enable fine-grained alignment with human preferences and values. Our proposed method, RRPO, can be applied to align generative models in both language and multimodal settings. By facilitating more precise alignment, RRPO has the potential to improve the safety, reliability, and usability of these models for real-world usage.

\section{Qualitative Results}\label{supsec:rrpo_qualitative}

\begin{figure}[h]
    \centering
    \includegraphics[width=\linewidth]{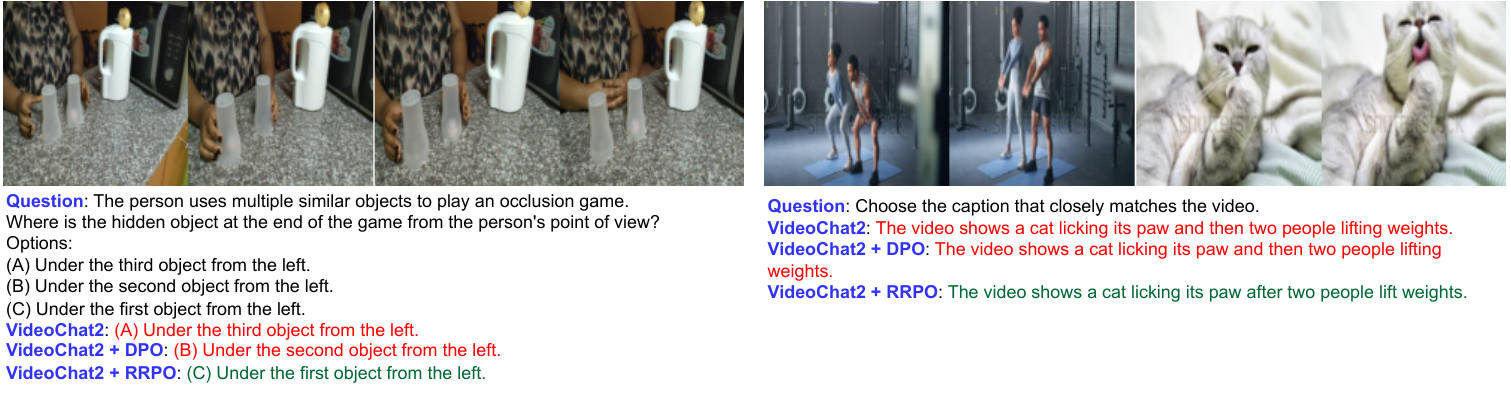}\\
    \includegraphics[width=\linewidth]{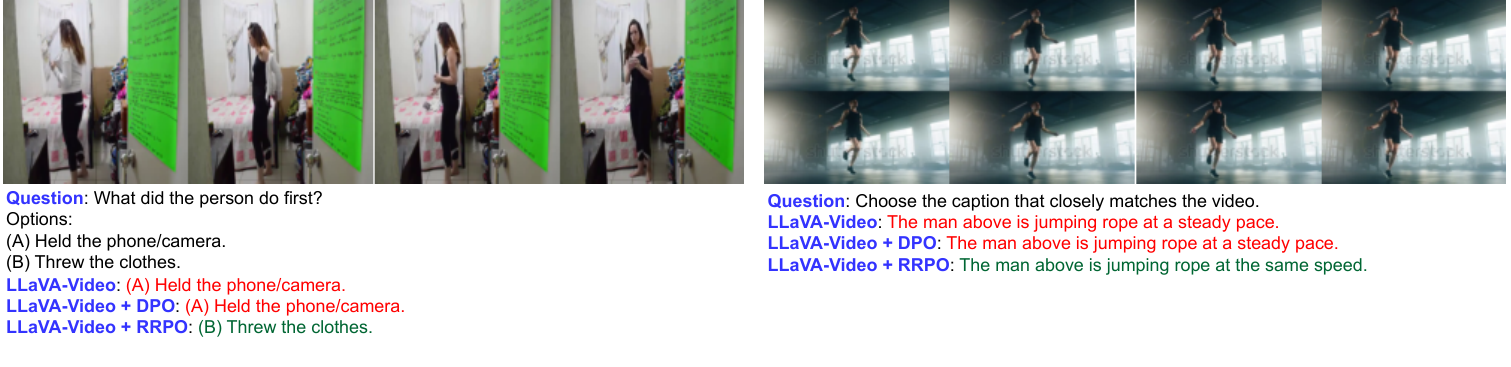}\\
    \includegraphics[width=\linewidth]{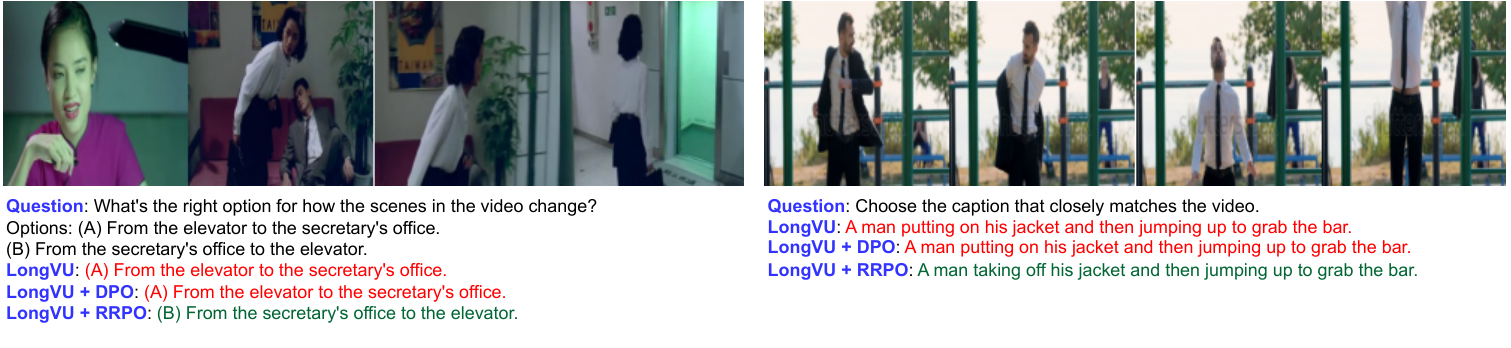}\\
    \caption{Qualitative examples on fine-grained temporal understanding tasks.}
    \label{fig:rrpo_fg}
\end{figure}

\begin{figure}[h]
    \centering
    \includegraphics[width=\linewidth]{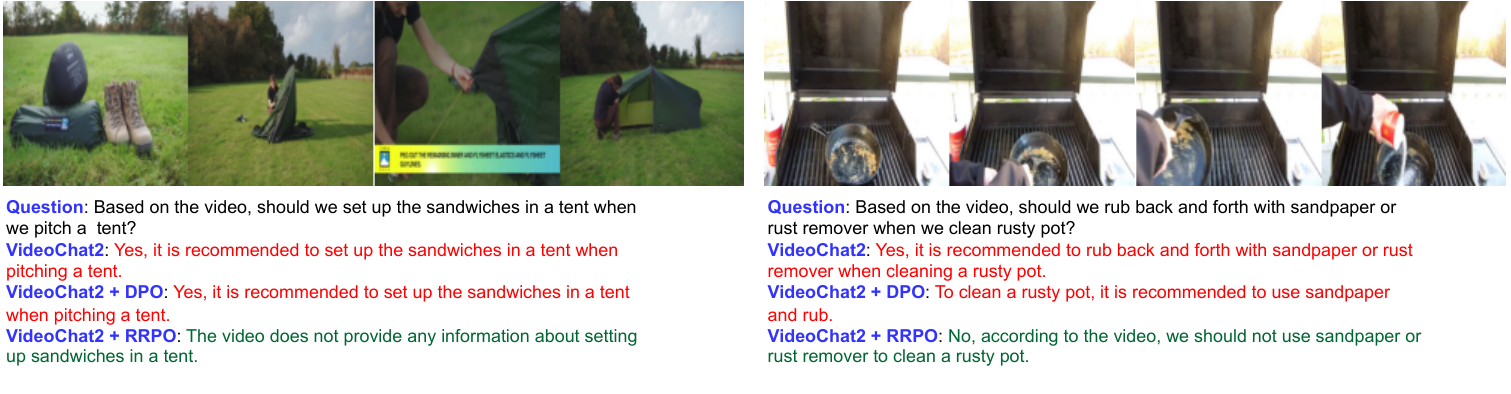}\\
    \includegraphics[width=\linewidth]{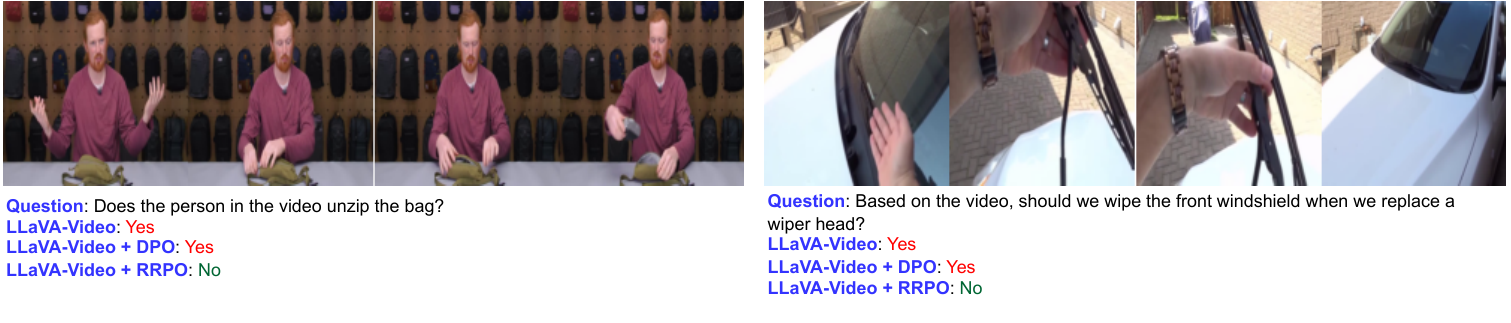}\\
    \includegraphics[width=\linewidth]{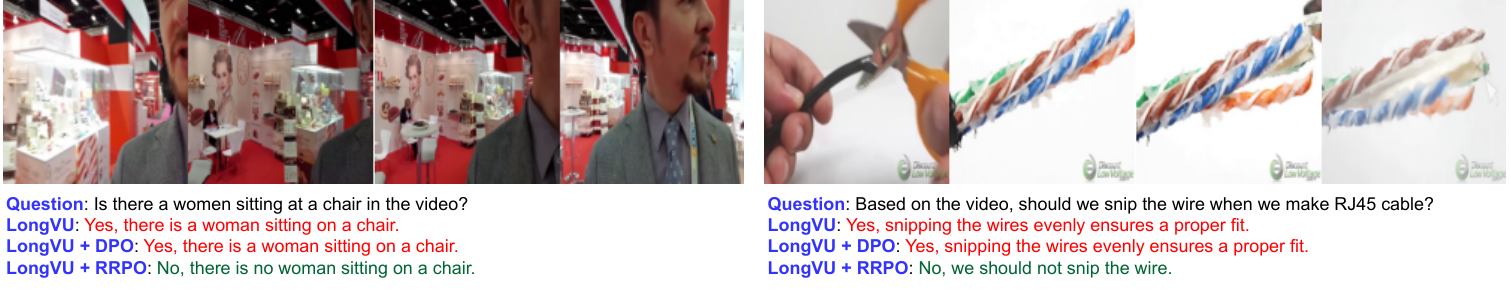}\\    
    \caption{Qualitative examples on video hallucination tasks.}
    \label{fig:rrpo_hal}
\end{figure}

\begin{figure}[h]
    \centering
    \includegraphics[width=\linewidth]{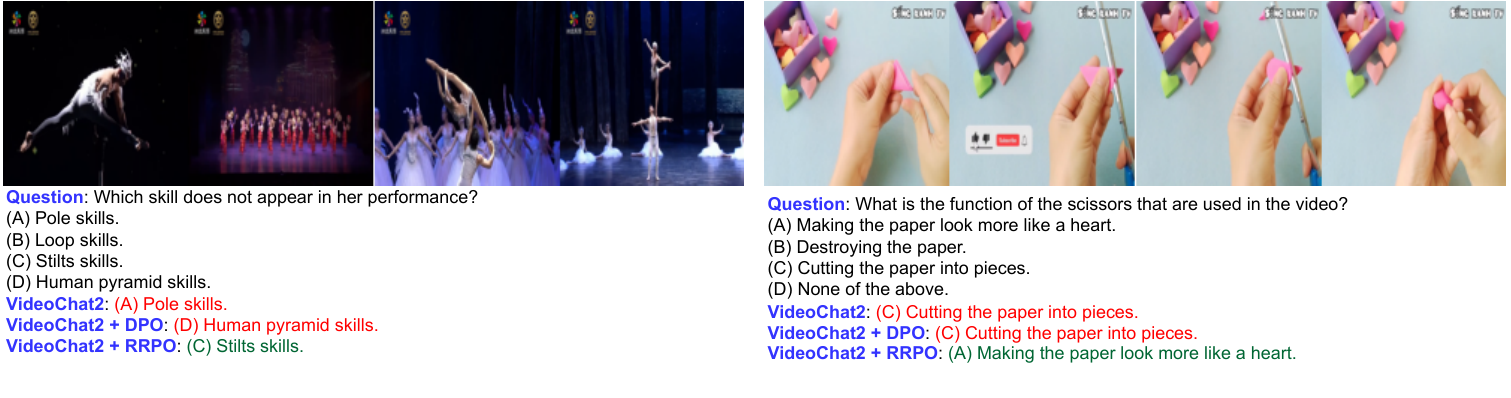}\\
    \includegraphics[width=\linewidth]{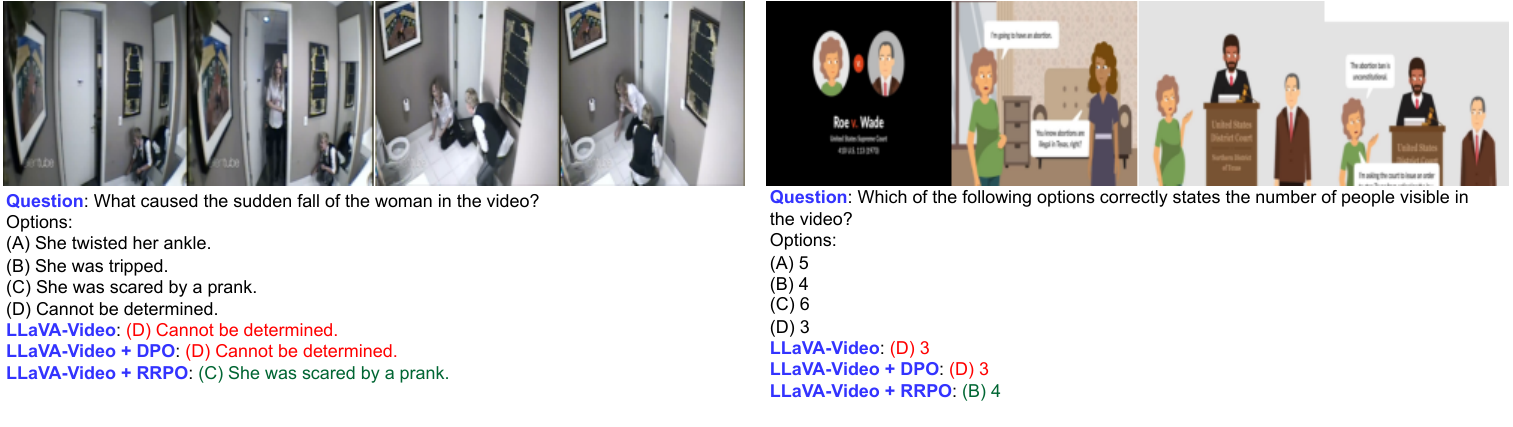}\\
    \includegraphics[width=\linewidth]{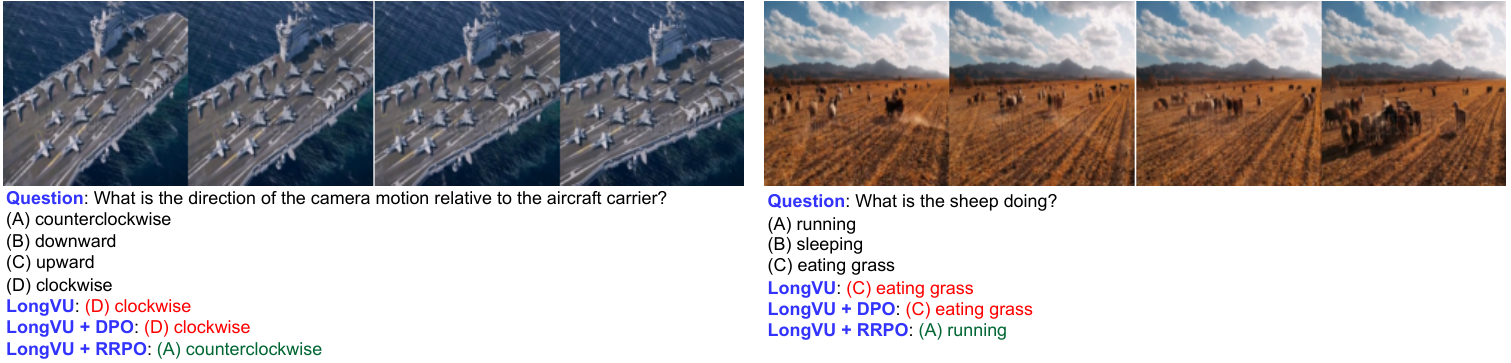}\\
    \caption{Qualitative examples on comprehensive video understanding tasks (short).}
    \label{fig:rrpo_comp_short}
\end{figure}

\begin{figure}[h]
    \centering
    \includegraphics[width=\linewidth]{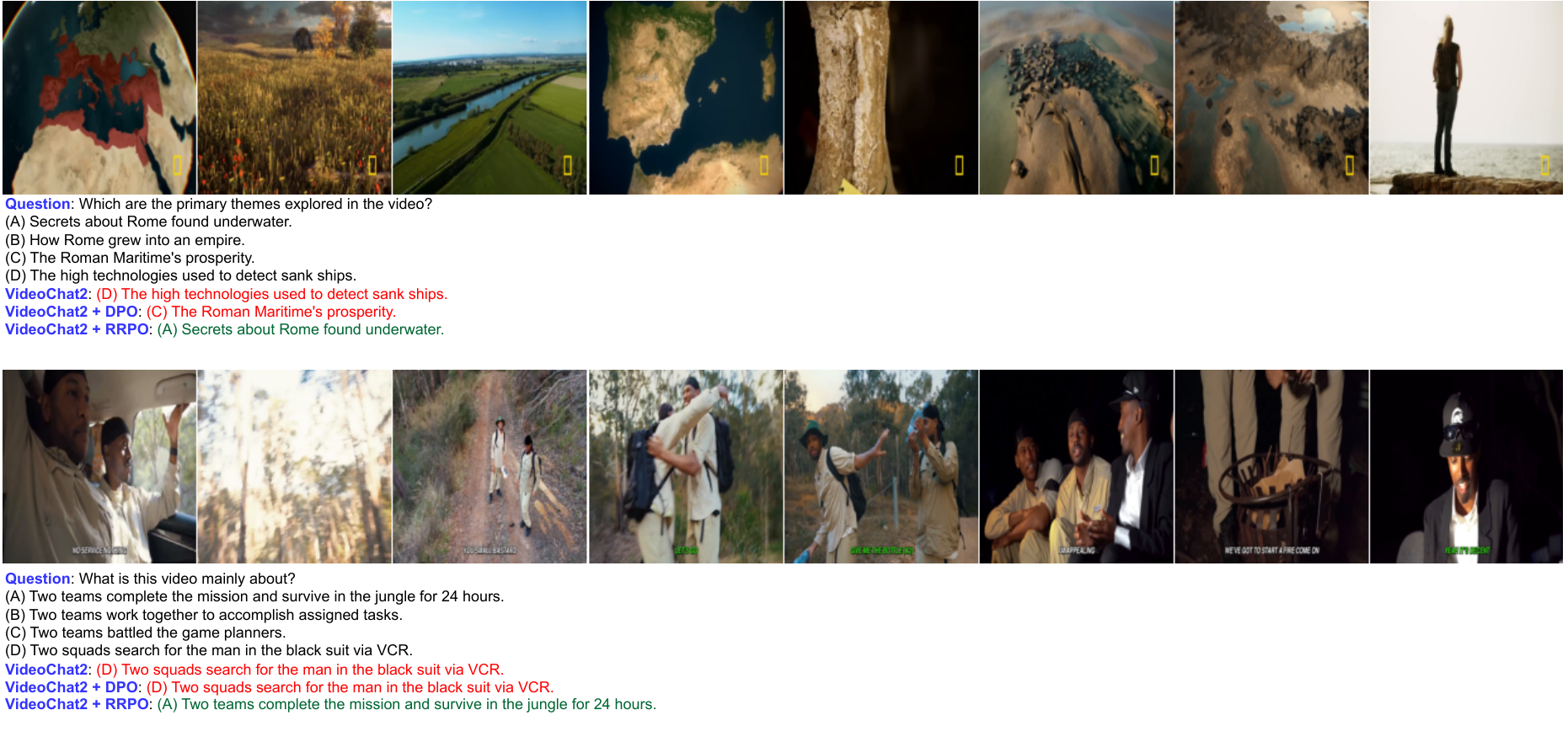}\\
    \includegraphics[width=\linewidth]{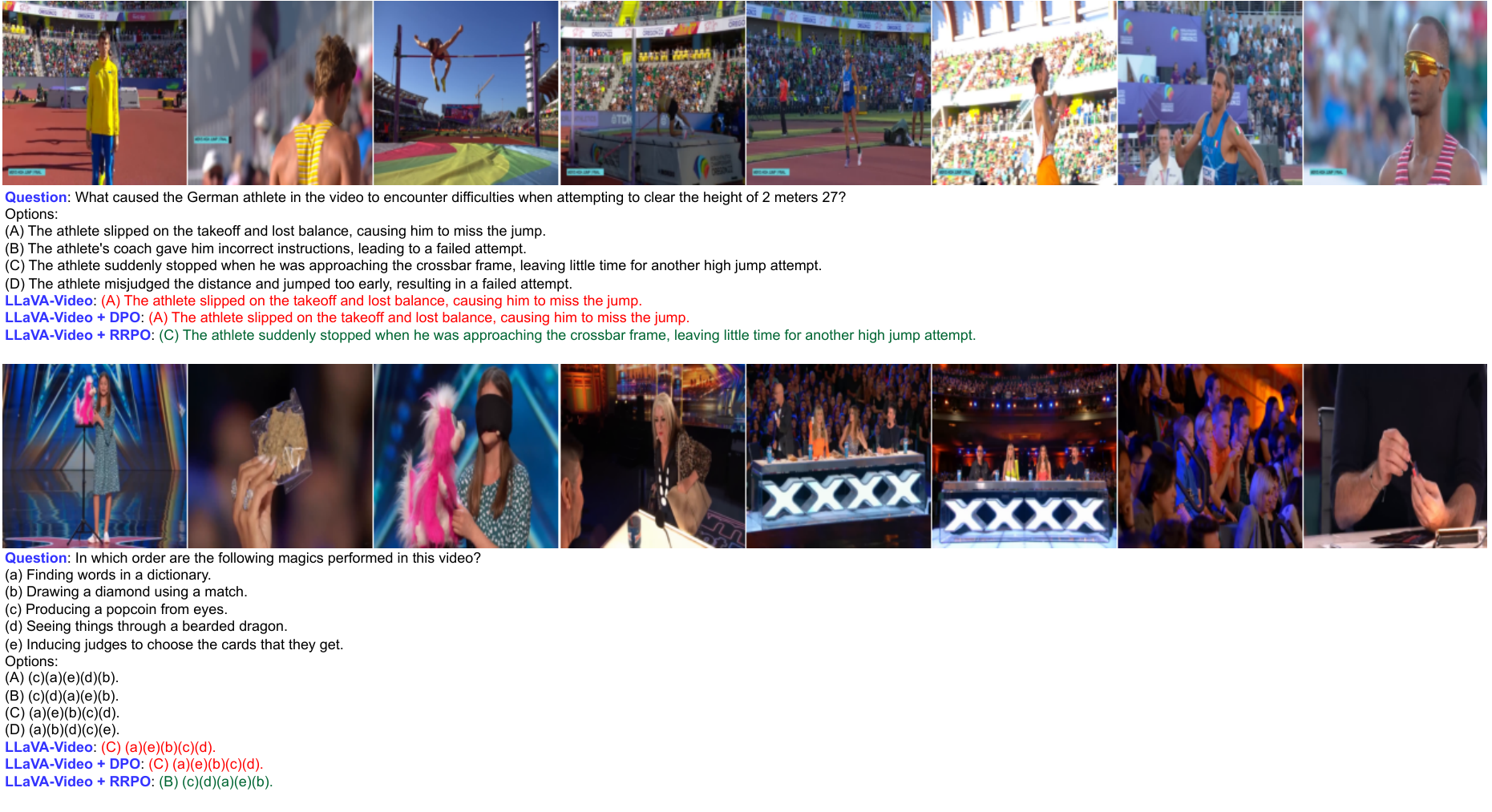}\\
    \includegraphics[width=\linewidth]{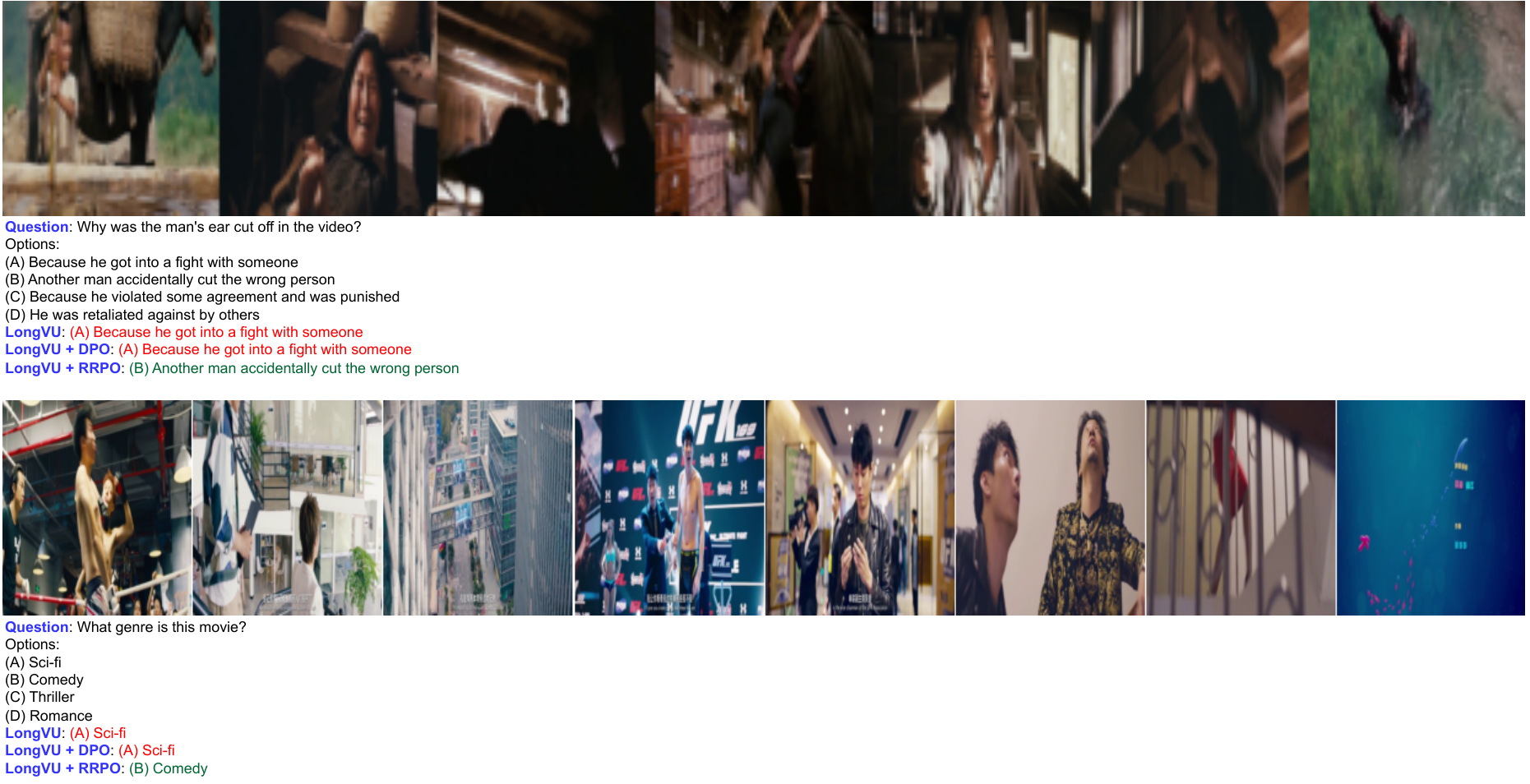}\\
    \caption{Qualitative examples on comprehensive video understanding tasks (long).}
    \label{fig:rrpo_comp_long}
\end{figure}

\begin{figure}[h]
    \centering
    \includegraphics[width=0.7\linewidth]{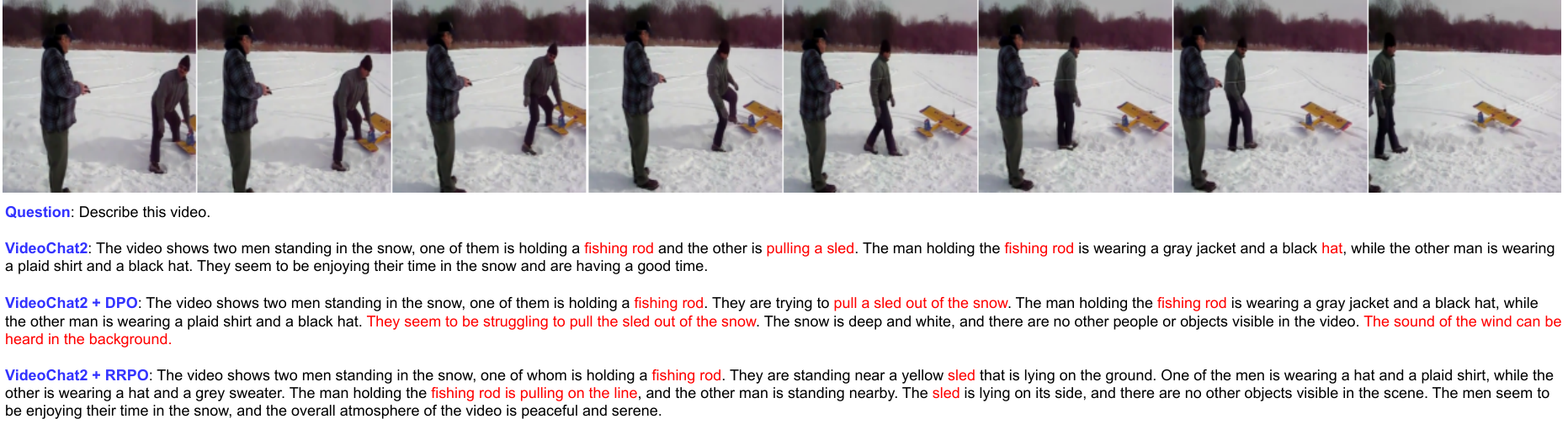}
    \includegraphics[width=0.7\linewidth]{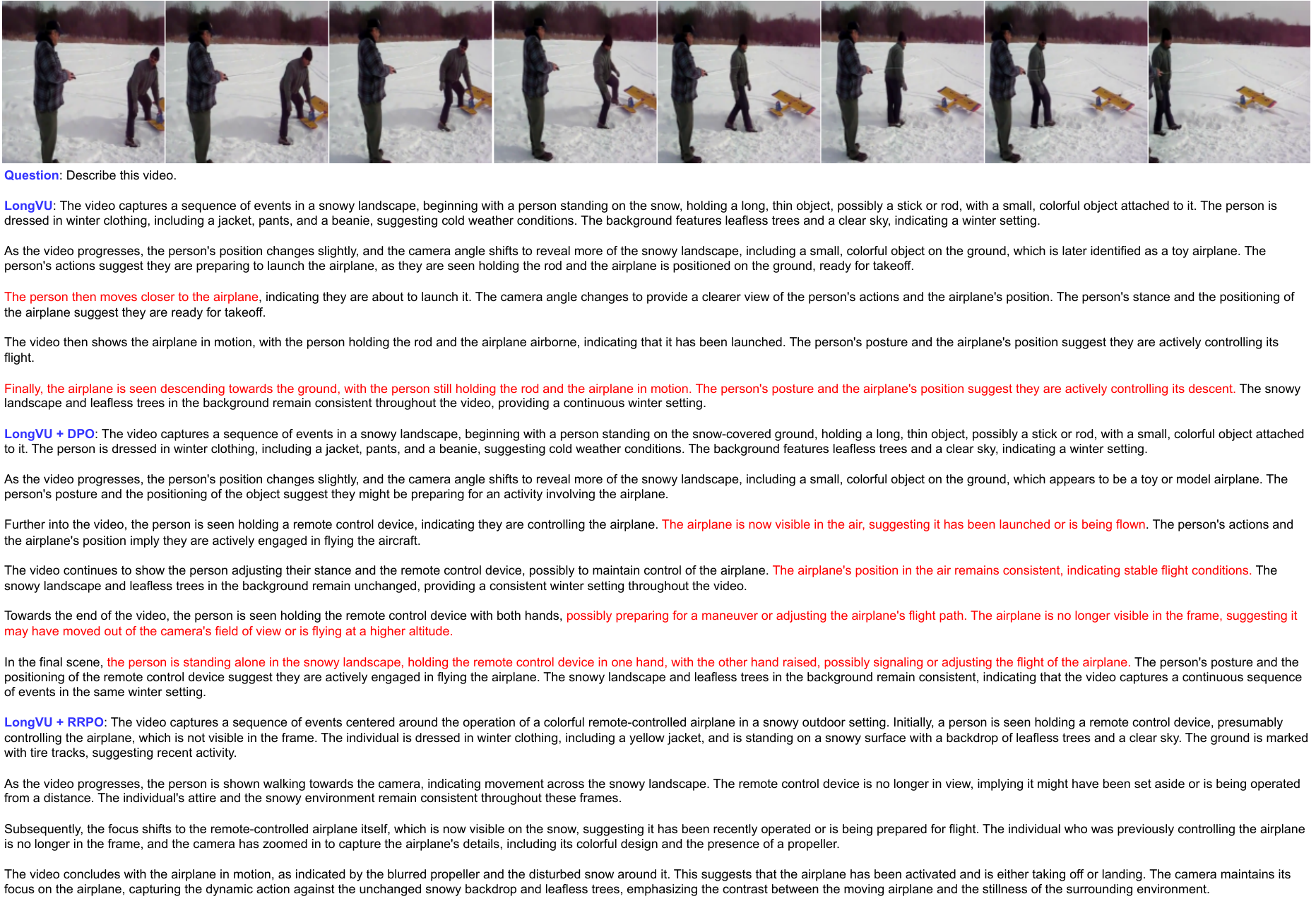}
    \includegraphics[width=0.7\linewidth]{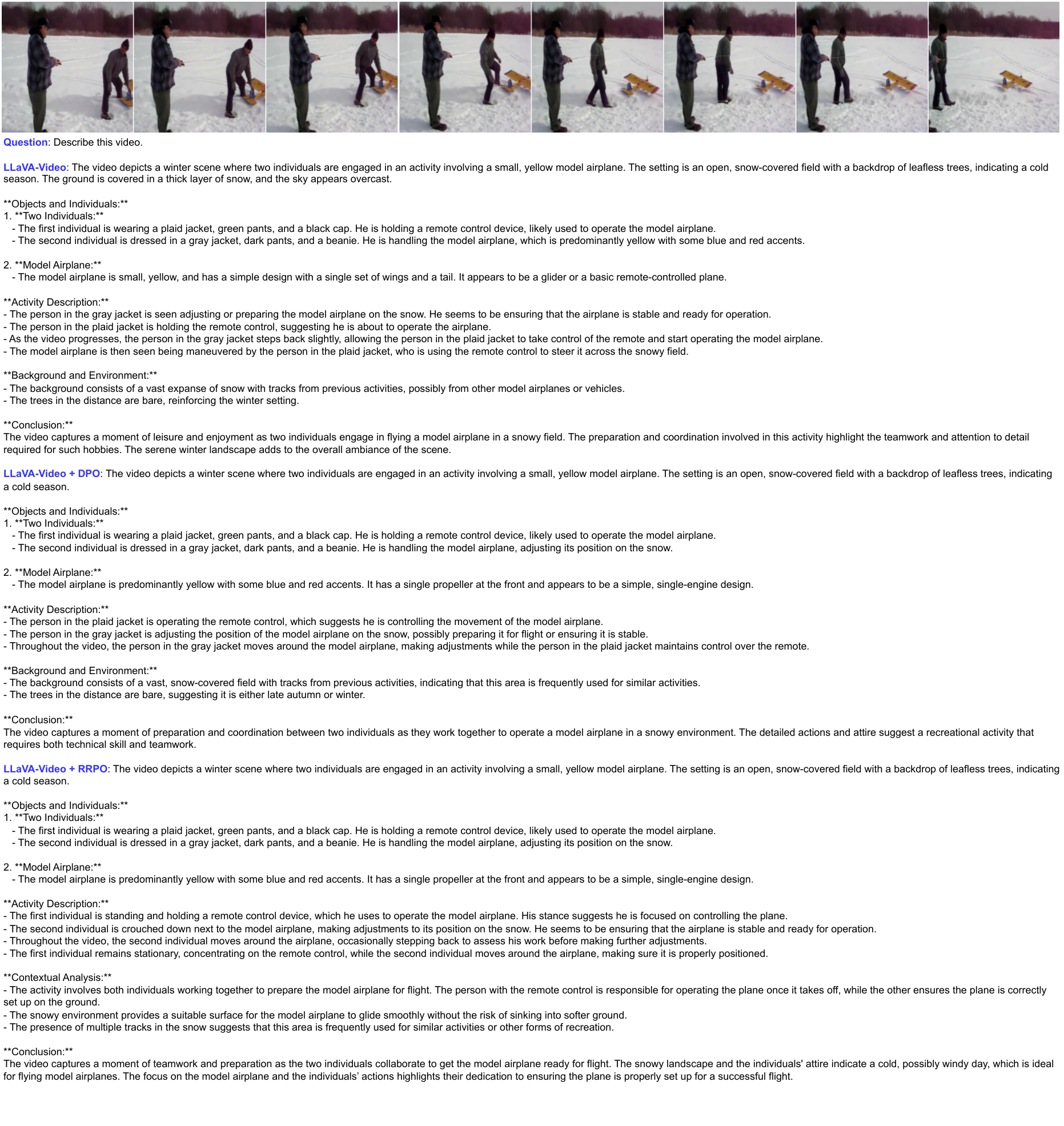}
    \caption{Building on the capabilities of the base models, we observe that RRPO-aligned models may still make mistakes or exhibit hallucinations in detailed video description tasks. For instance, VideoChat2 continues to display similar hallucinations after both DPO and RRPO training, as seen in the base model. In contrast, for LongVU, while both the base and DPO-aligned models hallucinate, the RRPO-aligned variant avoids such errors. Finally, in the case of LLaVA-Video, the RRPO-aligned model retains the base model’s reliable behavior, as neither exhibits hallucinations.}
    \label{fig:rrpo_failed_detailed}    
    
\end{figure}

\begin{figure}[h]
    \centering
    \includegraphics[width=\linewidth]{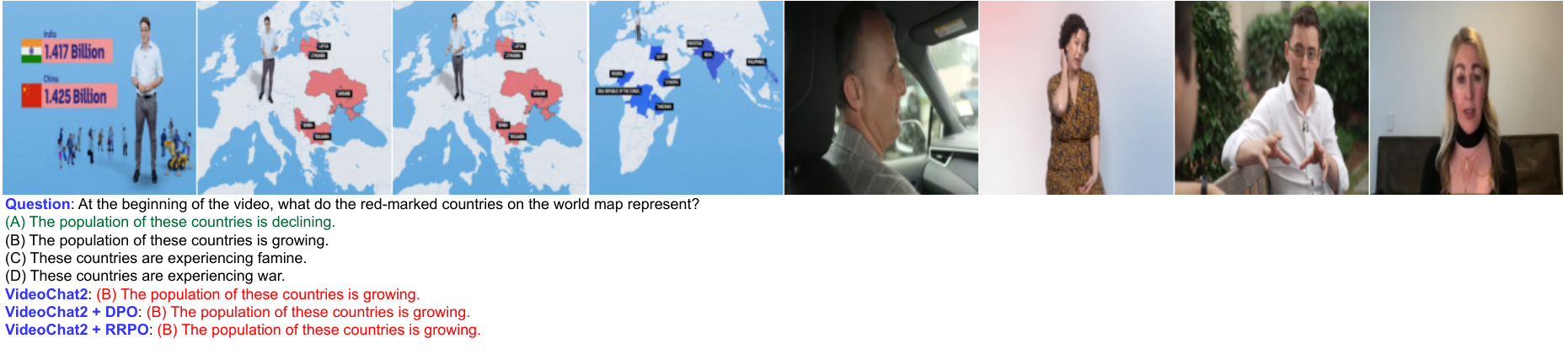}\\
    \includegraphics[width=\linewidth]{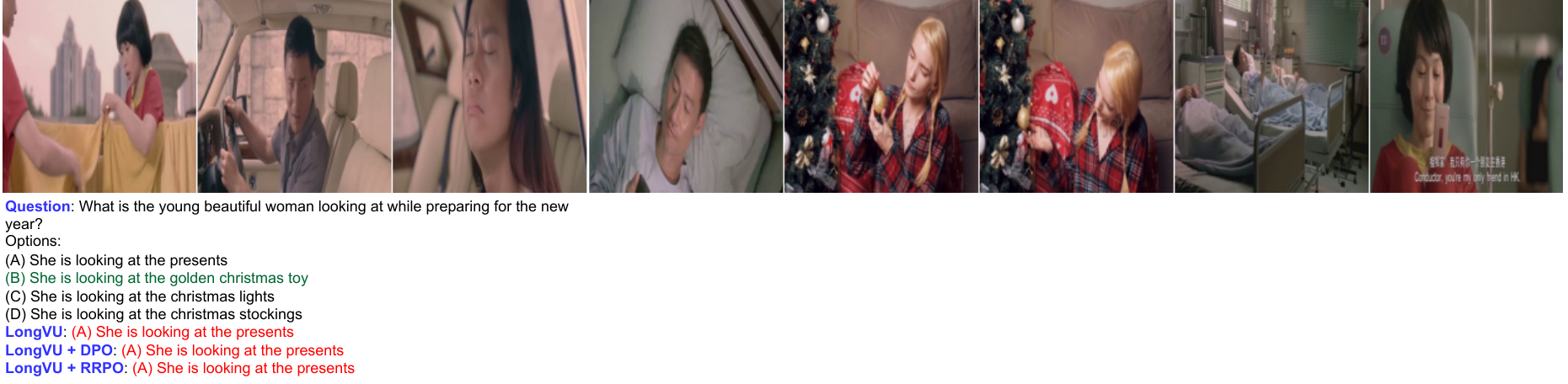}\\
    \includegraphics[width=\linewidth]{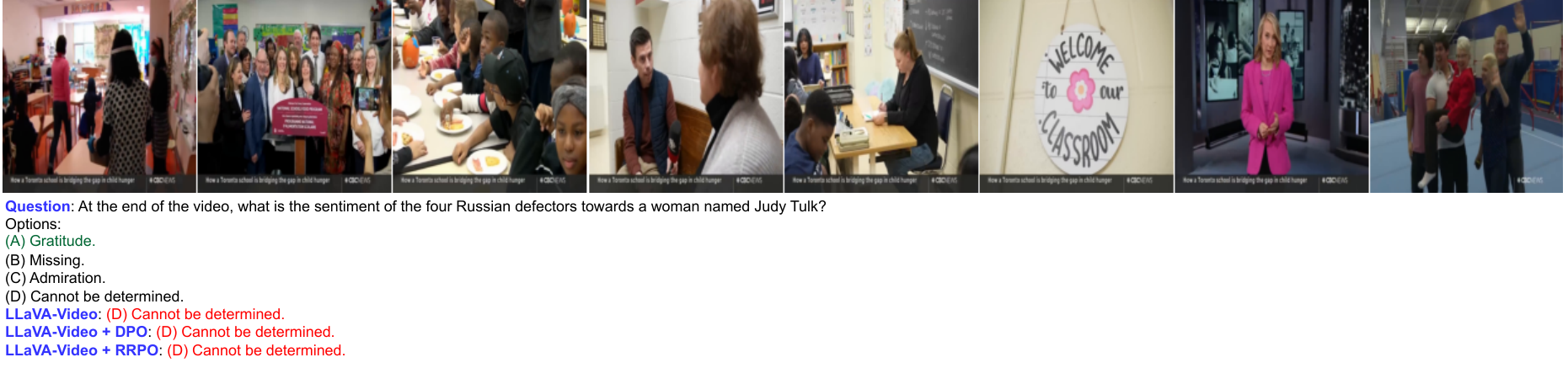}\\
    \caption{We also observe that RRPO-aligned models may still exhibit limitations in long video understanding tasks, primarily due to architectural constraints of the base model in processing extended frame sequences, as well as computational constraints during RRPO training that limit the use of long video inputs.}
    \label{fig:rrpo_failed_long}   
\end{figure}

We present several examples in Figures \ref{fig:rrpo_fg} - \ref{fig:rrpo_comp_long} highlighting the effectiveness of RRPO over the base model and other preference optimization methods (e.g., DPO) in diverse video understanding tasks. We also present some failure cases in Figures \ref{fig:rrpo_failed_detailed} - \ref{fig:rrpo_failed_long}.


\clearpage
\newpage
\section*{NeurIPS Paper Checklist}

\begin{enumerate}

\item {\bf Claims}
    \item[] Question: Do the main claims made in the abstract and introduction accurately reflect the paper's contributions and scope?
    \item[] Answer: \answerYes{} 
    \item[] Justification: The contributions of this paper are mentioned in the abstract and introduction.

\item {\bf Limitations}
    \item[] Question: Does the paper discuss the limitations of the work performed by the authors?
    \item[] Answer: \answerYes{} 
    \item[] Justification: The limitations are discussed.

\item {\bf Theory assumptions and proofs}
    \item[] Question: For each theoretical result, does the paper provide the full set of assumptions and a complete (and correct) proof?
    \item[] Answer: \answerYes{} 
    \item[] Justification: We present the mathematical derivation of our theoretical claims.  

    \item {\bf Experimental result reproducibility}
    \item[] Question: Does the paper fully disclose all the information needed to reproduce the main experimental results of the paper to the extent that it affects the main claims and/or conclusions of the paper (regardless of whether the code and data are provided or not)?
    \item[] Answer: \answerYes{} 
    \item[] Justification: The experimental results are fully reproducible, and sufficient details are shared in the paper. Moreover, we release the code and data through an anonymized repository during the review process for reproducibility. 

\item {\bf Open access to data and code}
    \item[] Question: Does the paper provide open access to the data and code, with sufficient instructions to faithfully reproduce the main experimental results, as described in supplemental material?
    \item[] Answer: \answerYes{} 
    \item[] Justification: The data and code are shared through an anonymized repository during the review process for reproducibility. 

\item {\bf Experimental setting/details}
    \item[] Question: Does the paper specify all the training and test details (e.g., data splits, hyperparameters, how they were chosen, type of optimizer, etc.) necessary to understand the results?
    \item[] Answer: \answerYes{} 
    \item[] Justification: The necessary details for the experimental settings are shared in the paper.

\item {\bf Experiment statistical significance}
    \item[] Question: Does the paper report error bars suitably and correctly defined or other appropriate information about the statistical significance of the experiments?
    \item[] Answer: \answerYes{} 
    \item[] Justification: 
    We test for statistical significance and consider performance improvements from the fine-tuned model over the base model statistically significant at the $95\%$ confidence level, based on Standard Error.

\item {\bf Experiments compute resources}
    \item[] Question: For each experiment, does the paper provide sufficient information on the computer resources (type of compute workers, memory, time of execution) needed to reproduce the experiments?
    \item[] Answer: \answerYes{} 
    \item[] Justification: The details of the computation requirements are discussed in the paper.
    
\item {\bf Code of ethics}
    \item[] Question: Does the research conducted in the paper conform, in every respect, with the NeurIPS Code of Ethics \url{https://neurips.cc/public/EthicsGuidelines}?
    \item[] Answer: \answerYes{} 
    \item[] Justification: The research conducted in the paper conforms, in every respect, with the NeurIPS Code of Ethics.

\item {\bf Broader impacts}
    \item[] Question: Does the paper discuss both potential positive societal impacts and negative societal impacts of the work performed?
    \item[] Answer: \answerYes{} 
    \item[] Justification: The Broader impact of this work is discussed in the paper.
    
\item {\bf Safeguards}
    \item[] Question: Does the paper describe safeguards that have been put in place for responsible release of data or models that have a high risk for misuse (e.g., pretrained language models, image generators, or scraped datasets)?
    \item[] Answer: \answerYes{} 
    \item[] Justification: This work involves finetuning publicly available, off-the-shelf large video-language models (LVLMs) to enhance their performance across diverse video understanding tasks. The base LVLMs were originally trained on carefully filtered public datasets and are already widely adopted in the research community. Moreover, our finetuning process uses open-source, carefully curated training data. We do not anticipate this process introducing any new safety concerns.

\item {\bf Licenses for existing assets}
    \item[] Question: Are the creators or original owners of assets (e.g., code, data, models), used in the paper, properly credited and are the license and terms of use explicitly mentioned and properly respected?
    \item[] Answer: \answerYes{} 
    \item[] Justification: Licenses for existing assets used in this work are mentioned in the paper.

\item {\bf New assets}
    \item[] Question: Are new assets introduced in the paper well documented and is the documentation provided alongside the assets?
    \item[] Answer: \answerYes{} 
    \item[] Justification: Yes, we introduce code, model, and data in this paper, which are available via an anonymous link during the review period and will be publicly released afterward. We also provide sufficient details of dataset/code/model as part of our submission throughout the main paper and the appendix.

\item {\bf Crowdsourcing and research with human subjects}
    \item[] Question: For crowdsourcing experiments and research with human subjects, does the paper include the full text of instructions given to participants and screenshots, if applicable, as well as details about compensation (if any)? 
    \item[] Answer: \answerNA{} 
    \item[] Justification: This work does not involve crowdsourcing nor research with human subjects.

\item {\bf Institutional review board (IRB) approvals or equivalent for research with human subjects}
    \item[] Question: Does the paper describe potential risks incurred by study participants, whether such risks were disclosed to the subjects, and whether Institutional Review Board (IRB) approvals (or an equivalent approval/review based on the requirements of your country or institution) were obtained?
    \item[] Answer: \answerNA{} 
    \item[] Justification: This work does not involve crowdsourcing nor research with human subjects.

\item {\bf Declaration of LLM usage}
    \item[] Question: Does the paper describe the usage of LLMs if it is an important, original, or non-standard component of the core methods in this research? Note that if the LLM is used only for writing, editing, or formatting purposes and does not impact the core methodology, scientific rigorousness, or originality of the research, declaration is not required.
    \item[] Answer: \answerNA{} 
    \item[] Justification: This work does not involve the use of LLMs in any way that affects the core methodology, scientific rigor, or originality of the research.

\end{enumerate}

\end{document}